\documentclass[twoside,11pt]{article}
\usepackage{jmlr2e}

\usepackage{algorithm}
\usepackage{algorithmic}
\usepackage{paralist,amsmath, amssymb}
\usepackage{microtype}
\usepackage{graphicx}
\usepackage{subfigure}
\usepackage{epstopdf}
\usepackage{booktabs} 
\def \A {\mathcal{A}}
\def \B {\mathcal{B}}
\def \R {\mathbb{R}}

\def \K {\mathcal{K}}
\def \F {\mathcal{F}}
\def \E {\mathbb{E}}
\def \U {\mathcal{U}}

\def \x {\mathbf{x}}
\def \z {\mathbf{z}}
\def \f {\tilde{f}}

\def \g {\mathbf{g}}
\def \v {\mathbf{v}}
\def \u {\mathbf{u}}
\def \y {\mathbf{y}}
\DeclareMathOperator*{\ii}{in}
\DeclareMathOperator*{\oo}{out}

\DeclareMathOperator*{\argmin}{argmin}
\DeclareMathOperator*{\argmax}{argmax}

\newtheorem{thm}{Theorem}
\newtheorem{lem}{Lemma}
\newtheorem{myDef}{Definition}
\newtheorem{assum}{Assumption}

\begin{document}

\title{Projection-free Online Learning with Arbitrary Delays}

\author{\name Yuanyu Wan\email wanyy@zju.edu.cn\\ 
\addr School of Software Technology, Zhejiang University, Ningbo 315048, China\\
       \name Yibo Wang \email wangyb@lamda.nju.edu.cn \\
\addr National Key Laboratory for Novel Software Technology, Nanjing University, Nanjing 210023, China\\
       \name Chang Yao\email changy@zju.edu.cn\\ 
\addr School of Software Technology, Zhejiang University, Ningbo 315048, China\\
       \name Wei-Wei Tu \email tuweiwei@4paradigm.com\\
\addr 4Paradigm Inc., Beijing 10000, China\\
       \name Lijun Zhang \email zhanglj@lamda.nju.edu.cn \\
\addr National Key Laboratory for Novel Software Technology, Nanjing University, Nanjing 210023, China
       }
\editor{}

\maketitle

\begin{abstract}
Projection-free online learning, which eschews the projection operation via less expensive computations such as linear optimization (LO), has received much interest recently due to its efficiency in handling high-dimensional problems with complex constraints. However, previous studies assume that any queried gradient is revealed immediately, which may not hold in practice and limits their applications. To address this limitation, we generalize the online Frank-Wolfe (OFW) algorithm and the online smooth projection-free (OSPF) algorithm, which are state-of-the-art LO-based projection-free online algorithms for non-smooth and smooth functions respectively, into a delayed setting where queried gradients can be delayed by arbitrary rounds. Specifically, the main idea of our generalized OFW is to perform an update similar to the original OFW after receiving any delayed gradient, and play the latest decision for each round. Moreover, the essential change on OSPF is to replace the sum of queried gradients, which is originally utilized in each update, with the sum of available gradients. Despite their simplicities, our novel analysis shows that under a relatively large amount of delay, the generalized OFW and OSPF enjoy the same regret bound as OFW and OSPF in the non-delayed setting, respectively.
\end{abstract}

\section{Introduction}
Online convex optimization (OCO) has become a leading paradigm for online learning due to its capability to model various problems from diverse domains such as online routing and online collaborative filtering \citep{Hazan2016}. In general, it is formulated as a structured repeated game between a player and an adversary. In each round $t$, the player first chooses a decision $\x_t$ from a convex constraint set $\K\subseteq\mathbb{R}^n$, where $n$ is the dimensionality. Then, the adversary selects a convex function $f_t(\x):\K\mapsto\mathbb{R}$, and the player suffers a loss $f_t(\x_t)$. The player aims to choose decisions such that the regret \[R(T)=\sum_{t=1}^Tf_t(\x_t)-\min_{\x\in\K}\sum_{t=1}^Tf_t(\x)\] is sublinear in the number of total rounds $T$. Online gradient descent (OGD) is a standard method for OCO, which enjoys an $O(\sqrt{T})$ regret bound for convex losses \citep{Zinkevich2003} and an $O(\log T)$ regret bound for strongly convex losses \citep{Hazan_2007}. Unfortunately, it needs to compute a projection onto the decision set to ensure the feasibility of each decision, which is computationally expensive for complex constraint sets \citep{Hazan2012}. 

Motivated by this computational issue, there has been a surge of research interest in developing projection-free OCO algorithms \citep{Hazan2012,Garber16,kevy_smooth,Wan-AAAI-2021-C,Garber-AISTATS21,Garber22,Zak_SC22,Zakaria21,Garber23}, which  eschews the projection operation via less expensive computations such as linear optimization (LO). For example, in the problem of online collaborative filtering, the constraint set consists of matrices with a bounded trace norm, and the linear optimization step is at least an order of magnitude faster than the projection operation \citep{Hazan2012}. However, these studies assume that the gradient $\nabla f_t(\x_t)$ is revealed immediately after making the decision $\x_t$, which is not necessarily satisfied in reality. For example, in the mentioned online collaborative filtering, the decision is a prediction of a user-item rating matrix, and the corresponding gradient depends on the true rating of a user on a item, which may not be decided by the user immediately. 

To address the above limitation, we investigate projection-free OCO in a more practical setting, in which the gradient $\nabla f_t(\x_t)$ arrives at the end of round $t+d_t-1$, and $d_t\geq 1$ denotes an arbitrary delay. It is well-known that in the non-delayed setting, the online Frank-Wolfe (OFW) algorithm \citep{Hazan2012,Wan-AAAI-2021-C,Garber-AISTATS21} and the online smooth projection-free (OSPF) algorithm \citep{Hazan20} are state-of-the-art LO-based projection-free algorithms for OCO with non-smooth and smooth losses, respectively. Specifically, in the non-smooth case, OFW can attain an $O(T^{3/4})$ regret bound in general and an $O(T^{2/3})$ regret bound when losses are strongly convex. If losses are smooth, OSPF can attain an $O(T^{2/3})$ regret bound without the strong convexity. Thus, it is natural to extend these two algorithms and their regret bounds into the setting with arbitrary delays.

One potential way for achieving this goal is to combine OFW and OSPF with an existing black-box technique for converting any traditional OCO algorithm into this delayed setting \citep{Joulani13}. To be precise, this black-box technique is to pool independent instances of a traditional algorithm $\A$,
each of which acts as a learner in the non-delayed setting over a subsequence of rounds. In each round, a single instance will be taken out from the pool, which makes a decision and then waits for its feedback before rejoining the pool. If the pool is empty, a new instance of $\A$ will be added to it. By using this approach, we are able to attain a regret bound of $dR(T/d)$ if $\A$ has the $R(T)$ regret over $T$ rounds, where $d$ is the maximum delay \citep{Joulani13}. As a result, combining this black-box technique with OFW and OSPF will attain $O(d^{1/4}T^{3/4})$, $O(d^{1/3}T^{2/3})$, and $O(d^{1/3}T^{2/3})$ regret bounds for convex, strongly convex, and smooth losses, respectively. These results magnify the regret bounds of OFW and OSPF in the non-delayed setting by a coefficient depending the delay. Thus, it is worthy to ask whether the effect of delay can be further reduced.
  
In this paper, we give an affirmative answer by respectively developing a better variant for OFW and OSPF, which are robust to a relatively large amount of delay. Different from the black-box technique that needs to maintain multiple instances of OFW and OSPF \citep{Joulani13}, our methods are much more simple and efficient. Specifically, the main idea of our variant of OFW, namely delayed OFW, is to update the decision similar to OFW after receiving any delayed gradient, and plays the latest decision for each round. Moreover, our variant of OSPF, namely delayed OSPF, only needs to replace the sum of queried gradients, which is originally utilized in each update, with the sum of available gradients. Our main theoretical contributions are summarized as follows. 
\begin{itemize}
  \item First, we prove that our delayed OFW attains an $O(T^{3/4}+\bar{d}T^{1/4})$ regret bound for convex losses, where $\bar{d}$ is the average delay, which matches the $O(T^{3/4})$ regret bound in the non-delayed setting as long as $\bar{d}$ does not exceed $O(\sqrt{T})$. 
  \item Second, we prove that our delayed OFW attains an $O(T^{2/3}+d\log T)$ regret bound for strongly convex losses, which matches the $O(T^{2/3})$ regret bound in the non-delayed setting as long as $d$ does not exceed $O(T^{2/3}/\log T)$.
  \item Third, we prove that our delayed OSPF attains an $O(T^{2/3}+dT^{1/3})$ regret bound for smooth losses, which matches the $O(T^{2/3})$ regret bound in the non-delayed setting, as long as $d$ does not exceed $O(T^{1/3})$.
\end{itemize}
Therefore, our regret bounds are strictly better than those achieved by combining the black-box technique \citep{Joulani13} with OFW and OSPF, when the term involving the delay in them is not dominant. 

The rest of the paper is structured as follows. In Section \ref{sec2}, we briefly review related work. Section \ref{sec3} presents our algorithms and their main theoretical guarantees. The detailed proofs for these results are provided in Section \ref{sec4}. 
Finally, we conclude this paper and discuss the future work in Section \ref{sec5}. We defer some additional results of our delayed OFW, which only hold in the special case with strongly convex sets, to Appendix \ref{sec-app1}. 
Note that a preliminary version of this paper was presented at the 36th Conference on Neural Information Processing Systems in 2022 \citep{NeurIPS22-Wan}.  In this paper, we have significantly enriched
the preliminary version by adding the following extensions.
\begin{itemize}
  \item We have significantly simplified the analysis of delayed OFW for convex losses and strongly convex losses. See the proofs of Theorems \ref{thm0} and \ref{thm1}.
  \item We improve the regret bound of delayed OFW for convex losses from $O(T^{3/4}+dT^{1/4})$ to $O(T^{3/4}+\bar{d}T^{1/4})$ via a more
careful analysis.
  \item We provide a new algorithm, namely delayed OSPF, which can exploit the smoothness of losses to improve the regret of delayed OFW for convex losses.
  \item We provide additional results of delayed OFW, which hold in the special case with strongly convex sets.
  \end{itemize}

\section{Related Work}
\label{sec2}
In this section, we briefly review related work on projection-free OCO algorithms, and OCO under delayed feedback.

\subsection{Projection-free OCO Algorithms}
The seminal work of \citet{Hazan2012} proposes the first projection-free algorithm for OCO, namely OFW, by extending the classical Frank-Wolfe algorithm \citep{FW-56,Revist_FW} from offline optimization into online optimization. Instead of performing the projection operation, OFW only utilizes a LO step to select a feasible decision, and can attain a sublinear regret bound of $O(T^{3/4})$ for the general OCO. If losses are strongly convex, \citet{Wan-AAAI-2021-C} prove that OFW can also be utilized to achieve an $O(T^{2/3})$ regret bound. The same regret bound is concurrently established by \citet{Garber-AISTATS21} in a similar way. If the decision set is strongly convex, \citet{Wan-AAAI-2021-C} show that OFW can further attain an $O(T^{2/3})$ regret bound for convex losses and an $O(\sqrt{T})$ regret bound for strongly convex losses. Moreover, if the decision set is polyhedral, \citet{Garber16} propose a specific variant of OFW, which is able to achieve an $O(\sqrt{T})$ regret bound for convex losses and an $O(\log T)$ regret bound for strongly convex losses.

To further exploit the smoothness of losses, \citet{Hazan20} propose a randomized LO-based projection-free algorithm, namely OSPF, which is an efficient variant of the classical follow the perturbed leader algorithm \citep{Kalai05} and can achieve a regret bound of $O(T^{2/3})$ for convex and smooth losses. \citet{Zak_SC22} proposes a LO-based projection-free algorithm based on novel techniques in parameter-free online learning \citep{zak_COLT20}, which can attain an $O(\sqrt{T\log T})$ regret bound for convex functions over strongly convex sets, and thus is better than OFW in the same case. Very recently, \citet{Garber23} develop a deterministic LO-based projection-free algorithm for exponentially concave and smooth functions, and establish an $O(T^{2/3})$ regret bound.

Besides LO, some other efficient computations including the membership and separation oracles over the decision set have been utilized to develop projection-free algorithms over smooth sets \citep{kevy_smooth} and general sets \citep{Zakaria21,Garber22}, which can achieve $O(\sqrt{T})$ and $O(\log T)$ regret bounds for convex losses and strongly convex losses, respectively. Although these bounds are better than the regret bounds achieved by LO-based projection-free algorithms, as discussed in previous studies \citep{Zak_SC22,Zakaria21}, there exist many decision sets in which LO is much more efficient than the membership and separation oracles. Moreover, we note that projection-free algorithms for other OCO scenarios such as the distributed setting \citep{wenpeng17,Wan-ICML-2020,Wan-JMLR22,Wang_AAAI23}, the bandit setting \citep{chen19,Garber19,Garber-AISTATS21}, and the dynamic environments \citep{Kalhan2021,Wan-AAAI-2021-B,Garber22,Lu_ALT23,Wan-2023-OFW} have also been proposed. 

However, despite this great flourish of research on projection-free OCO algorithms, the practical
problem of delayed feedback has not been considered. In this paper, we take the first step towards understanding the effect of delays on the performance of projection-free OCO algorithms by investigating OFW and OSPF with arbitrary delays.

\subsection{OCO under Delayed Feedback}
The starting point for studies on OCO under delayed feedback is the seminal work of \citet{Weinberger02_TIT}, which focuses on a special case with a fixed delay $d^\prime$, i.e., the feedback for each decision $\x_t$ is received at the end of round $t+d^\prime-1$. They propose a technique that can convert any traditional OCO algorithm for the non-delayed setting into this delayed setting. Specifically, their technique is to run $d^\prime$ instances of a traditional OCO algorithm, where each instance is used every $d^\prime$ rounds, which is feasible because the feedback for any decision $\x_t$ must be available for selecting the decision $\x_{t+d^\prime}$. If the traditional OCO algorithm enjoys a regret bound of $R(T)$ in the non-delayed setting, they proved that this technique achieves a regret bound of $d^\prime R(T/d^\prime)$. As a result, by combining with OGD, the regret bound of this technique is $O(\sqrt{d^\prime T})$ for convex losses and $O(d^\prime\log T)$ for strongly convex losses. However, this technique needs to run $d^\prime$ instances of OGD, which requires more storage and computational resources than the original OGD. To address this problem, \citet{Langford09} study the same setting, and show that simply performing each gradient descent step in the original OGD with a delayed gradient can also achieve the $O(\sqrt{d^\prime T})$ regret bound for convex losses and the $O(d^\prime\log T)$ regret bound for strongly convex losses. Moreover, \citet{Shamir17} combine the fixedly delayed OCO with a local permutation setting in which the order of losses can be modified by a distance of $M\geq d^\prime$, and propose an algorithm with the $O(\sqrt{T}(1+\sqrt{(d^\prime)^2/M}))$ regret for convex losses. This result implies that the permutation on the order of losses can be utilized to alleviate the effect of the fixed delay.

Different the above studies that only consider the fixed delay, \citet{Joulani13} extend the technique in \citet{Weinberger02_TIT} to handle a more general setting, where each feedback is delayed by arbitrary rounds. For this setting, their technique is able to attain a regret bound of $dR(T/d)$ by combining with a traditional OCO algorithm with $R(T)$ regret, where $d$ is the maximum delay. Note that although this technique can also convert OFW and OSPF to the delayed setting, it can only attain $O(d^{1/4}T^{3/4})$, $O(d^{1/3}T^{2/3})$, and $O(d^{1/3}T^{2/3})$ regret bounds for convex losses, strongly convex losses, and smooth losses, which cannot match regret bounds of OFW and OSPF in the non-delayed setting as long as $d$ is larger than $\Omega(1)$. Additionally, similar to the technique in \citet{Weinberger02_TIT}, it also needs to run multiple instances of a traditional OCO algorithm, which could be prohibitively resource-intensive. 
For this reasons, many studies have proposed delayed OCO algorithms, which only require the same storage and computational resources as in the non-delayed setting. 

Specifically, \citet{Quanrud15} propose a delayed variant of OGD, which simply performs a gradient descent update according to the sum of gradients arrived in each round, and establish an $O(\sqrt{\bar{d}T})$ regret bound for convex losses, where $\bar{d}$ is the average delay. \citet{Joulani16} develop adaptive algorithms that enjoy data-dependent regret with arbitrary delays, which could be tighter than $O(\sqrt{\bar{d}T})$ for sparse data. The main idea is to update similar to existing adaptive algorithms in the non-delayed setting \citep{McMahan10,Duchi2011} after receiving the full information of any function, and always play the latest decision. \citet{ICML21_delay} extend optimistic algorithms \citep{Rakhlin2013,Joulani2017} into the delayed setting, which can improve the regret by utilizing “hints” about future losses. \citet{Li_AISTATS19} develop a delayed bandit convex optimization (BCO) algorithm, which enjoys the $O(\sqrt{\bar{d}T})$ regret bound while only querying the function value at $(n+1)$ points per round, instead of the gradient. \citet{ICML20_Mertikopoulos} consider a special case with $d_t=o(t^c)$ for some known $c<1$, and propose another delayed BCO algorithm, which only queries the function value at one point per round, and attains a regret bound of $\tilde{O}(T^{3/4}+T^{2/3+c/3})$.\footnote{We use the $\tilde{O}$ notation to hide constant factors as well as polylogarithmic factors in $T$.} Recently, \citet{DOGD-SC} further show that by using an appropriate step size, the algorithm of \citet{Quanrud15} enjoys a regret bound of $O(d\log T)$ for strongly convex losses, where $d$ is the maximum delay. However, these studies do not consider projection-free algorithms.

\section{Main Results}
\label{sec3}
In this section, we first introduce necessary preliminaries including the problem setting, definitions, and assumptions. Then, we present our delayed OFW for convex losses and strongly convex losses, respectively. Finally, we introduce our delayed OSPF for smooth losses.

\subsection{Preliminaries}
We consider the problem of OCO with arbitrary delays \citep{Joulani13,Quanrud15}. Similar to the standard OCO, in each round $t=1,\dots,T$, the player first chooses a decision $\x_t$ from the decision set $\K$, and then the adversary selects a convex function $f_t(\x)$. However, different from the standard OCO, the gradient $\nabla f_t(\x_t)$ is revealed at the end of round $t+d_t-1$, where $d_t\geq 1$ denotes an arbitrary delay. 
As a result, the player actually receives gradients $\{\nabla f_k(\x_k)|k\in\F_t\}$ at the end of round $t$, where \[\F_t=\{k|k+d_k-1=t\}.\] 
Then, we recall the definitions of smooth functions and strongly convex functions \citep{optimization_book,Boyd04}.
\begin{myDef}
\label{def1}
A function $f(\x):\K\to\mathbb{R}$ is called $\alpha$-smooth over $\K$ if for all $\x,\mathbf{y}\in \K$, it holds that
\[\|\nabla f(\x)-\nabla f(\y)\|_2\leq\alpha\|\x-\y\|_2.\]
\end{myDef}
\begin{myDef}
\label{def2}
A function $f(\x):\K\to\mathbb{R}$ is called $\beta$-strongly convex over $\K$ if for all $\x,\mathbf{y}\in \K$, it holds that \[f(\mathbf{y})\geq f(\x)+\langle\nabla f(\x),\mathbf{y}-\x\rangle+\frac{\beta}{2}\|\mathbf{y}-\x\|_2^2.\]
\end{myDef}
Finally, similar to previous studies about OCO \citep{Online:suvery,Hazan2016}, we introduce two assumptions, which are required by both delayed OFW and delayed OSPF.
\begin{assum}
\label{assum1}
The gradients of all loss functions are bounded by $G$, i.e., for any $\x\in\K$ and $t\in[T]$, it holds that \[\|\nabla f_t(\x)\|_2\leq G.\]
\end{assum}
\begin{assum}
\label{assum2}
The decision set $\K$ contains the origin $\mathbf{0}$, and its diameter is bounded by $D$, i.e., for any $\x,\y\in\K$, it holds that
\[\|\x-\y\|_2\leq D.\]
\end{assum}

\subsection{Delayed OFW for Convex Losses}
In the non-delayed setting, OFW for convex losses \citep{Hazan2012,Hazan2016} first
chooses an arbitrary $\x_1\in\K$, and then iteratively updates its decision by the following linear optimization step
\begin{equation}
\label{sec2-eq1}
\begin{split}
\v_t\in\argmin_{\x\in\K}\langle\nabla F_t(\x_t),\x\rangle,~\x_{t+1}=\x_t+\sigma_t(\v_t-\x_t)
\end{split}
\end{equation}
where $F_t(\x)$ is a surrogate loss defined as
\begin{equation}
\label{sec2-eq2}
\begin{split}
F_t(\x)=\eta\sum_{i=1}^t\langle\nabla f_i(\x_i),\x\rangle+\|\x-\x_1\|_2^2
\end{split}
\end{equation}
and $\eta,\sigma_t$ are two parameters. Notice that according to (\ref{sec2-eq2}), OFW for convex losses requires the gradient $\nabla f_t(\x_t)$ before making the decision $\x_{t+1}$. However, in the problem of OCO with arbitrary delays, this requirement is not necessarily satisfied, because the player actually receives gradients $\{\nabla f_k(\x_k)|k\in\F_t\}$ at each round $t$, which may not contain the gradient $\nabla f_t(\x_t)$. 

To address this limitation, our main idea is to update the decision similar to OFW for each received gradient, and play the latest decision for each round. 
In this way, there exist some intermediate decisions that are not really played. To facilitate presentations, we introduce an additional notation $\y_\tau$ to denote the $\tau$-th intermediate decision. Moreover, we denote the sum of $\tau$ received gradients by $\bar{\g}_\tau$. Initially, we choose an arbitrary vector $\y_1\in\K$ and set $\tau=1,\bar{\g}_0=\mathbf{0}$. At each round $t=1,\dots,T$, we play the latest decision
$\x_t=\y_{\tau}$ and query the gradient $\nabla f_t(\x_t)$. Then, we receive delayed gradients that are queried in a set of rounds $\F_t$, and perform the following steps for each received gradient. 
\begin{algorithm}[t]
\caption{Delayed OFW for Convex Losses}
\label{alg1}
\begin{algorithmic}[1]
\STATE \textbf{Input:} $\eta$
\STATE \textbf{Initialization:} choose an arbitrary vector $\y_1\in\K$ and set $\tau=1,\bar{\g}_0=\mathbf{0}$
\FOR{$t=1,2,\dots,T$}
\STATE Play $\x_t=\y_{\tau}$ and query $\nabla f_t(\x_t)$
\STATE Receive a set of delayed gradients $\{\nabla f_k(\x_k)|k\in \F_t\}$
\FOR{$k\in \F_t$}
\STATE Update $\bar{\g}_\tau=\bar{\g}_{\tau-1}+\nabla f_k(\x_k)$ and define $F_\tau(\y)=\eta\langle\bar{\g}_\tau,\y\rangle+\|\y-\y_1\|_2^2$
\STATE Compute $\v_\tau\in\argmin_{\y\in\K}\langle\nabla F_\tau(\y_\tau),\y\rangle$
\STATE Update $\y_{\tau+1}=\y_\tau+\sigma_\tau(\v_\tau-\y_\tau)$ with $\sigma_\tau$ in (\ref{sec3-eq3}) and set $\tau = \tau+1$
\ENDFOR
\ENDFOR
\end{algorithmic}
\end{algorithm}

Inspired by (\ref{sec2-eq2}) of the original OFW, for any $k\in\F_t$, we first compute $\bar{\g}_\tau=\bar{\g}_{\tau-1}+\nabla f_k(\x_k)$ and define
\begin{equation*}
\begin{split}
F_\tau(\y)=\eta\langle\bar{\g}_\tau,\y\rangle+\|\y-\y_1\|_2^2.
\end{split}
\end{equation*}
Then, similar to (\ref{sec2-eq1}) of the original OFW, we perform the following update 
\begin{equation*}
\begin{split}
\v_\tau\in\argmin_{\y\in\K}\langle\nabla F_\tau(\y_\tau),\y\rangle,~\y_{\tau+1}=\y_\tau+\sigma_\tau(\v_\tau-\y_\tau)
\end{split}
\end{equation*}
where $\sigma_\tau$ is a parameter. Following \citet{Wan-AAAI-2021-C}, it is set by a line search rule
\begin{equation}
\label{sec3-eq3}
\sigma_\tau=\argmin_{\sigma\in[0,1]}\langle\sigma(\v_\tau-\y_\tau),\nabla F_\tau(\y_\tau)\rangle+\sigma^2\|\v_\tau-\y_\tau\|_2^2.
\end{equation}
Finally, we update $\tau=\tau+1$ so that $\tau$ still indexes the latest intermediate decision. 

The detailed procedures are summarized in Algorithm \ref{alg1}, which is named as delayed OFW for convex losses. Notice that $\bar{d}=(1/T)\sum_{t=1}^Td_t$ and ${d}=\max\{d_1,\dots,d_T\}$ according to their definitions. Then, we establish the following theorem with respect to the regret of Algorithm \ref{alg1}.
\begin{thm}
\label{thm0}
For any $\x^\ast\in\K$, under Assumptions \ref{assum1} and \ref{assum2}, Algorithm \ref{alg1} with $\eta=\frac{D}{GT^{3/4}}$ has
\[\sum_{t=1}^Tf_t(\x_t)-\sum_{t=1}^Tf_t(\x^\ast)\leq7GDT^{3/4}+GD(\bar{d}+1)T^{1/4}.\]
\end{thm}
\begin{remark}
\emph{Theorem \ref{thm0} shows that without knowing the value of $\bar{d}$, our Algorithm \ref{alg1} can attain an $O(T^{3/4}+\bar{d}T^{1/4})$ regret bound for convex losses with arbitrary delays. This bound matches the $O(T^{3/4})$ regret bound of OFW for convex losses in the non-delayed setting \citep{Hazan2012,Hazan2016}, as long as the average delay $\bar{d}$ does not exceed $O(\sqrt{T})$. Moreover, it is better than the $O(d^{1/4}T^{3/4})$ regret bound achieved by combining the technique of \citet{Joulani13} and the $O(T^{3/4})$ regret bound of OFW for convex losses, as long as $\bar{d}$ does not exceed $O(T^{2/3})$.}
\end{remark}

\subsection{Delayed OFW for Strongly Convex Losses}
Before introducing our algorithm for strongly convex losses, we notice that in the non-delayed setting, to exploit the $\beta$-strongly convex property of losses, \citet{Wan-AAAI-2021-C} have refined the surrogate loss function in (\ref{sec2-eq1}) to
\begin{equation}
\label{sec2-eq3}
\begin{split}
F_t(\x)=\sum_{i=1}^t\left(\langle\nabla f_i(\x_i),\x\rangle+\frac{\beta}{2}\|\x-\x_i\|_2^2\right).
\end{split}
\end{equation} 
The main difference is that the regularization term in (\ref{sec2-eq3}) is about all historical decisions, instead of only the initial decision. 

Inspired by (\ref{sec2-eq3}), we first redefine $F_{\tau}(\y)$ in Algorithm \ref{alg1} to
\begin{equation*}
F_\tau(\y)= \langle\bar{\g}_\tau,\y\rangle+\sum_{i=1}^\tau\frac{\beta}{2}\|\y-\y_i\|_2^2.
\end{equation*}
Second, since $F_{\tau}(\y)$ is modified, we adjust the line search rule to
\begin{equation}
\label{sec3-eq5}
\sigma_\tau=\argmin_{\sigma\in[0,1]}\langle\sigma(\v_\tau-\y_\tau),\nabla F_\tau(\y_\tau)\rangle+\frac{\beta\tau\sigma^2}{2}\|\v_\tau-\y_\tau\|_2^2.
\end{equation}
The detailed procedures are summarized in Algorithm \ref{alg2}, which is named as delayed OFW for strongly convex losses. Then, we establish the following theorem about the regret of Algorithm \ref{alg2}.
\begin{algorithm}[t]
\caption{Delayed OFW for Strongly Convex Losses}
\label{alg2}
\begin{algorithmic}[1]
\STATE \textbf{Input:} $\beta$
\STATE \textbf{Initialization:} choose an arbitrary vector $\y_1\in\K$ and set $\tau=1,\bar{\g}_0=\mathbf{0}$
\FOR{$t=1,2,\dots,T$}
\STATE Play $\x_t=\y_{\tau}$ and query $\nabla f_t(\x_t)$
\STATE Receive a set of delayed gradients $\{\nabla f_k(\x_k)|k\in \F_t\}$
\FOR{$k\in \F_t$}
\STATE Update $\bar{\g}_\tau=\bar{\g}_{\tau-1}+\nabla f_k(\x_k)$ and define $F_\tau(\y)= \langle\bar{\g}_\tau,\y\rangle+\sum_{i=1}^\tau\frac{\beta}{2}\|\y-\y_i\|_2^2$
\STATE Compute $\v_\tau\in\argmin_{\y\in\K}\langle\nabla F_\tau(\y_\tau),\y\rangle$
\STATE Update $\y_{\tau+1}=\y_\tau+\sigma_\tau(\v_\tau-\y_\tau)$ with $\sigma_\tau$ in (\ref{sec3-eq5}) and set $\tau = \tau+1$
\ENDFOR
\ENDFOR
\end{algorithmic}
\end{algorithm}

\begin{thm}
\label{thm1}
Suppose all losses are $\beta$-strongly convex and Assumptions \ref{assum1} and \ref{assum2} hold. For any $\x^\ast\in\K$, Algorithm \ref{alg2} has
\begin{align*}
\sum_{t=1}^Tf_t(\x_t)-\sum_{t=1}^Tf_t(\x^\ast)\leq&\frac{6\sqrt{2}C_2(C_1+2C_2)T^{2/3}}{\beta}+\frac{2C_2^2\ln T}{\beta}+C_1D\\
&+(C_1+C_2)\left(3{d}D+\frac{4\sqrt{2}{d}C_2}{\beta}+\frac{2{d}C_1}{\beta}\ln T\right)
\end{align*}
where $C_1=G+\beta D$ and $C_2=G+2\beta D$.
\end{thm}
\begin{remark}
\emph{Theorem \ref{thm1} shows that our Algorithm \ref{alg2} can attain an $O(T^{2/3}+d\log T)$ regret bound for strongly convex losses with arbitrary delays. First, this bound matches the $O(T^{2/3})$ regret bound of OFW for strongly convex losses in the non-delayed setting \citep{Wan-AAAI-2021-C}, as long as $d$ does not exceed $O(T^{2/3}/\log T)$.
Second, as long as $d$ does not exceed $O(T^{3/4}/\log T)$, this bound is smaller than $O(T^{3/4})$, and thus is better than the $O(T^{3/4}+\bar{d}T^{1/4})$ regret bound in Theorem \ref{thm0}, which is established by only using the convexity condition. 
 Third, it is better than the $O(d^{1/3}T^{2/3})$ regret bound achieved by combining the technique of \citet{Joulani13} and the $O(T^{2/3})$ regret bound of OFW for strongly convex losses, as long as $d$ does not exceed $O(T/(\log T)^{3/2})$.}
\end{remark}

\subsection{Delayed OSPF for Smooth Losses}
Notice that similar to the original OFW, one limitation of our delayed OFW is that it cannot utilize the smoothness of losses. In the non-delayed setting, OSPF \citep{Hazan20} can address this limitation. Specifically, OSPF divides the total $T$ rounds into $T/K$ blocks, and utilizes the same decision $\x_m$ for rounds in each block $m\in[T/K]$, where $K$ is the block size.\footnote{Without loss of generality, $T/K$ is assumed to be an integer.} To select the the decision $\x_m$, in the beginning of each block $m\in[T/K]$, it first computes $K$ intermediate decisions as
\begin{equation}
\label{rule-OSPF}
\x_m^j=\argmax_{\x\in\K}\left\langle-\sum_{i=1}^{m-1}\sum_{t=(i-1)K+1}^{iK}\nabla f_t(\x_i)+\frac{\v_{(m-1)K+j}}{\delta},\x\right\rangle
\end{equation}
for $j=1,\dots,K$, where $\sum_{i=1}^{m-1}\sum_{t=(i-1)K+1}^{iK}\nabla f_t(\x_i)$ is the sum of gradients queried before block $m$, $\v_{(m-1)K+1},\dots,\v_{(m-1)K+K}$ is uniformly sampled from the unit ball $\B=\left\{\x\in\mathbb{R}^n|\|\x\|_2\leq1\right\}$, and $\delta$ is a parameter. Then, the decision $\x_m$ is simply computed by averaging these intermediate decisions, i.e.,
\[\x_m=\frac{1}{K}\sum_{j=1}^K\x_{m}^j.\]
However, in the problem of OCO with arbitrary delays, there may exist some gradients that are queried before block $m$, but not received in the beginning of this block, which makes (\ref{rule-OSPF}) infeasible.

To address this limitation, we simply replace the sum of gradients queried before block $m$ with the sum of all available gradients in the beginning of this block. To this end, we modify (\ref{rule-OSPF}) to
\begin{equation}
\label{rule-DOSPF}
\x_m^j=\argmax_{\x\in\K}\left\langle-\sum_{i=1}^{m-1}\sum_{t=(i-1)K+1}^{iK}\sum_{k\in\F_{t}}\nabla f_k(\x_{m_k})+\frac{\v_{(m-1)K+j}}{\delta},\x\right\rangle
\end{equation}
where $m_k=\lceil k/K\rceil$ denotes the block index of round $k$. By using $\hat{\g}_t$ to denote the sum of available gradients at the end of round $t$, the term $\sum_{i=1}^{m-1}\sum_{t=(i-1)K+1}^{iK}\sum_{k\in\F_{t}}\nabla f_k(\x_{m_k})$ in (\ref{rule-DOSPF}) can be simplified to $\hat{\g}_{(m-1)K}$. Then, the detailed procedures are summarized in Algorithm \ref{alg3}, which is named as delayed OSFW for smooth losses. Despite the simplicity of our modification, in the following theorem, we demonstrate that Algorithm \ref{alg3} is also robust to a relatively large amount of delay.
\begin{algorithm}[t]
\caption{Delayed OSPF for Smooth Losses}
\label{alg3}
\begin{algorithmic}[1]
\STATE \textbf{Input:} $\delta,K$
\STATE \textbf{Initialization:} set $\hat{\g}_0=\mathbf{0}$
\FOR{$m=1,2,\dots,T/K$}
\STATE Sample $\v_{(m-1)K+j}\sim\B$ uniformly for $j=1,\dots,K$, where $\B=\left\{\x\in\mathbb{R}^n|\|\x\|_2\leq1\right\}$
\STATE Compute $\x_m^j=\argmax_{\x\in\K}\left\langle-\hat{\g}_{(m-1)K}+\frac{\v_{(m-1)K+j}}{\delta},\x\right\rangle$ and $\x_m=\frac{1}{K}\sum_{j=1}^K\x_{m}^j$
\FOR{$t=(m-1)K+1,(m-1)K+2,\dots,mK$}
\STATE Play $\x_m$ and query $\nabla f_t(\x_m)$
\STATE Receive a set of delayed gradients $\{\nabla f_k(\x_{m_k})|k\in \F_t\}$, where $m_k=\lceil k/K\rceil$
\STATE Update $\hat{\g}_t=\hat{\g}_{t-1}+\sum_{k\in\F_t}\nabla f_k(\x_{m_k})$
\ENDFOR
\ENDFOR
\end{algorithmic}
\end{algorithm}

\begin{thm}
\label{thm1_ospf}
Suppose all losses are $\alpha$-smooth, Assumptions \ref{assum1} and \ref{assum2} hold, and the adversary is oblivious. For any $\x^\ast\in\K$, Algorithm \ref{alg3} with $\delta=\frac{2}{\sqrt{n}GT^{2/3}}$ and $K=T^{1/3}$ has
\[\E\left(\sum_{m=1}^{T/K}\sum_{t=(m-1)K+1}^{mK}\left(f_t(\x_m)-f_t(\x^\ast)\right)\right)\leq4\alpha D^2T^{2/3}+2\sqrt{n}{d}DGT^{1/3}+2\sqrt{n}DGT^{2/3}.\]
\end{thm}
\begin{remark}
\emph{First, we notice that similar to the original OSPF \citep{Hazan20}, our Algorithm \ref{alg3} needs to assume that the adversary is oblivious, i.e., all loss functions and delays are chosen beforehand. Second, Theorem \ref{thm1_ospf} shows that Algorithm \ref{alg3} can attain an $O(T^{2/3}+{d}T^{1/3})$ regret bound for smooth losses with arbitrary delays. As long as the maximum delay ${d}$ does not exceed $O(T^{1/3})$, this bound matches the $O(T^{2/3})$ regret bound of OSPF for smooth losses in the non-delayed setting. Third, as long as ${d}$ does not exceed $O(T^{5/12})$, this bound is smaller than $O(T^{3/4})$, and thus is better than the $O(T^{3/4}+\bar{d}T^{1/4})$ regret bound of delayed OFW for convex losses. Moreover, it is better than the $O(d^{1/3}T^{2/3})$ regret bound achieved by combining the technique of \citet{Joulani13} and the $O(T^{2/3})$ regret bound of OSPF for smooth losses, as long as $d$ does not exceed $O(\sqrt{T})$.}
\end{remark}

\section{Theoretical Analysis}
\label{sec4}
Notice that in the non-delayed setting, the regret of OFW and OSPF are worse than that of OGD due to the gap between the linear optimization step and the projection operation. Thus, in the delayed setting, it is natural to conjecture that both the linear optimization step and the delay will affect the regret of our algorithms. To keep the same regret bound as the original OFW and OSPF, our key insight is to prove that the combined effect of the linear optimization step and the delay can be additive, instead of being multiplicative, and the effect of the delay can be weaker than that of the linear optimization step for a relatively large amount of delay. In this following, we provide detailed proofs of all theorems. 
\subsection{Preliminaries for the Analysis of Algorithms \ref{alg1} and \ref{alg2}}
Before starting this proof, we introduce some necessary preliminaries. First, because of the effect of delays, there may exist some gradients that arrive after round $T$. Although our Algorithms \ref{alg1} and \ref{alg2} do not need to use these gradients, they are useful for the analysis. As a result, in our analysis, we further set $\x_t=\y_\tau$ and perform steps $5$ to $10$ of Algorithms \ref{alg1} and \ref{alg2} for any $t=T+1,\dots,T+{d}-1$. In this way, all gradients $\nabla f_1(\x_1),\nabla f_2(\x_2),\dots,\nabla f_T(\x_T)$ queried by these algorithms are utilized, which produces decisions \[\y_1,\y_2,\dots,\y_{T+1}.\] 
Then, let $\tau_t=1+\sum_{i=1}^{t-1}|\F_i|$ for any $t\in[T+{d}-1]$. It is not hard to verify that our Algorithms \ref{alg1} and \ref{alg2} ensure that \begin{equation}
\label{re_index0}
\x_t=\y_{\tau_t}
\end{equation} for any $t\in[T+{d}-1]$. 

To further facilitate the analysis, we denote the time-stamp of the $\tau$-th gradient used in the update of our Algorithms \ref{alg1} and \ref{alg2} by $c_\tau$. To help understanding, one can imagine that our Algorithms \ref{alg1} and \ref{alg2} also implement $c_\tau=k$ in their step $7$. 
By using this notation, $F_\tau(\y)$ defined in Algorithms \ref{alg1} and \ref{alg2} are respectively equivalent to
\begin{align}
F_\tau(\y)=&\eta\sum_{i=1}^\tau\langle\nabla f_{c_i}(\x_{c_i}),\y\rangle+\|\y-\y_1\|_2^2,\label{sec4-1-eq1}\\
F_\tau(\y)=&\sum_{i=1}^\tau\langle\nabla f_{c_i}(\x_{c_i}),\y\rangle+\sum_{i=1}^\tau\frac{\beta}{2}\|\y-\y_i\|_2^2. \label{sec4-1-eq2}
\end{align}

\subsection{Proof of Theorem \ref{thm0}} 
According to the convexity of $f_t(\x)$, we have
\begin{equation}
\label{thm0_eq1}
\begin{split}
\sum_{t=1}^Tf_t(\x_t)-\sum_{t=1}^Tf_t(\x^\ast)\leq&\sum_{t=1}^T\langle\nabla f_{t}(\x_{t}),\x_t-\x^\ast\rangle=\sum_{t=1}^T\langle\nabla f_{c_t}(\x_{c_t}),\x_{c_t}-\x^\ast\rangle\\
=&\sum_{t=1}^T\langle\nabla f_{c_t}(\x_{c_t}),\y_{t}-\x^\ast\rangle+\sum_{t=1}^T\langle\nabla f_{c_t}(\x_{c_t}),\y_{\tau_{c_t}}-\y_{t}\rangle\\
\leq&\sum_{t=1}^T\langle\nabla f_{c_t}(\x_{c_t}),\y_{t}-\x^\ast\rangle+\sum_{t=1}^TG\|\y_{\tau_{c_t}}-\y_{t}\|_2
\end{split}
\end{equation}
where the first equality is due to the fact that $c_1,\dots,c_T$ is a permutation of $1,\dots,T$, and the second equality is due to (\ref{re_index0}).

Then, we consider the term $\sum_{t=1}^TG\|\y_{\tau_{c_t}}-\y_{t}\|_2$ in (\ref{thm0_eq1}). By carefully analyzing the distance $\|\y_{\tau_{c_t}}-\y_{t}\|_2$, we establish the following lemma.
\begin{lem}
\label{c_c_lemma_term_A}
Let $\y_{t}^\ast=\argmin_{\y\in \K}F_{t-1}(\y)$ for any $t\in[T+1]$, where $F_t(\y)$ is defined in (\ref{sec4-1-eq1}). Suppose Assumptions \ref{assum1} and \ref{assum2} hold, and there exist some constants $\gamma>0$ and $\alpha$ such that $F_{t-1}(\y_{t})-F_{t-1}(\y_t^\ast)\leq \gamma T^{-\alpha}$ for any $t\in[T+1]$. Then, Algorithm \ref{alg1} ensures
\begin{equation*}
\begin{split}
\sum_{t=1}^T\|\y_{\tau_{c_t}}-\y_{t}\|_2
\leq&\eta G\bar{d}T+2\sqrt{\gamma}T^{1-\alpha/2}.
\end{split}
\end{equation*}
\end{lem}
Note that Lemma \ref{c_c_lemma_term_A} introduces an assumption about $\y_t$ and $F_{t-1}(\y)$. According to our Algorithm \ref{alg1}, $\y_t$ is actually generated by approximately minimizing $F_{t-1}(\y)$ with a linear optimization step. Therefore, inspired by the analysis of the original OFW \citep{Hazan2016}, we show that this assumption can be satisfied with $\gamma=4D^2$ and $\alpha=1/2$.
\begin{lem}
\label{thm0_lem1}
Let $\y_{t}^\ast=\argmin_{\y\in \K}F_{t-1}(\y)$ for any $t\in[T+1]$, where $F_t(\y)$ is defined in (\ref{sec4-1-eq1}). Under Assumptions \ref{assum1} and \ref{assum2}, for any $t\in[T+1]$, Algorithm \ref{alg1} with $\eta=\frac{D}{GT^{3/4}}$ has
\[F_{t-1}(\y_{t})-F_{t-1}(\y_t^\ast)\leq\frac{4D^2}{\sqrt{T}}.\]
\end{lem}
Then, by combining $\eta=\frac{D}{GT^{3/4}}$ with Lemmas \ref{c_c_lemma_term_A} and \ref{thm0_lem1}, we have
\begin{equation}
\label{eq_119_1}
\begin{split}
\sum_{t=1}^TG\|\y_{\tau_{c_t}}-\y_{t}\|_2
\leq&GD\bar{d}T^{1/4}+4GDT^{3/4}.
\end{split}
\end{equation}
Moreover, for the term $\sum_{t=1}^T\langle\nabla f_{c_t}(\x_{c_t}),\y_{t}-\x^\ast\rangle$ in (\ref{thm0_eq1}), inspired by the analysis of the original OFW \citep{Hazan2016}, we can establish the following lemma.
\begin{lem}
\label{thm0_lem0}
Under Assumptions \ref{assum1} and \ref{assum2}, for any $\x^\ast\in\K$, Algorithm \ref{alg1} with $\eta=\frac{D}{GT^{3/4}}$ ensures
\begin{equation*}
\begin{split}
\sum_{t=1}^{T}\langle\nabla f_{c_t}(\x_{c_t}),\y_{t}-\x^\ast\rangle\leq 3GDT^{3/4}+GDT^{1/4}.
\end{split}
\end{equation*}
\end{lem}
Finally, by combining (\ref{thm0_eq1}), (\ref{eq_119_1}), and Lemma \ref{thm0_lem0}, we complete this proof.

\subsection{Proof of Lemma \ref{c_c_lemma_term_A}}
We start this proof by introducing a nice property of strongly convex functions and a property of $\tau_{c_t}$. First, note that $F_t(\y)$ is $2$-strongly convex for any $t=0,\dots,T$, and \citet{Hazan2012} have proved that for any $\beta$-strongly convex function $f(\x)$ over $\K$ and any $\x\in\K$, it holds that
\begin{equation}
\label{cor_scvx}
\frac{\beta}{2}\|\x-\x^\ast\|_2^2\leq f(\x)-f(\x^\ast)
\end{equation}
where $\x^\ast=\argmin_{\x\in\K} f(\x)$.

Then, we also notice that \[\y_1,\dots,\y_{\tau_{c_t}}\] have already been generated before round $c_t$, and $\nabla f_{c_t}(\x_{c_t})$ can only be used to update $\y_t$ in round $c_t$ or later. Therefore, for any $t\in[T]$, it is easy to verify that
\begin{equation}
\label{short_eq2}
\tau_{c_t}\leq t\leq T.
\end{equation}
Now, we proceed to bound $\sum_{t=1}^T\|\y_{\tau_{c_t}}-\y_{t}\|_2$. It is easy to verify that
\begin{equation}
\label{short_eq1}
\begin{split}
\sum_{t=1}^T\|\y_{\tau_{c_t}}-\y_{t}\|_2\leq&\sum_{t=1}^T\left(\|\y_{\tau_{c_t}}-\y_{\tau_{c_t}}^\ast\|_2+\|\y_{\tau_{c_t}}^\ast-\y_{t}^\ast\|_2+\|\y_t^\ast-\y_{t}\|_2\right).
\end{split}
\end{equation}
Because of (\ref{cor_scvx}), for any $t\in[T+1]$, we have
\begin{equation}
\label{thm0_eq5}
\begin{split}
\|\y_{t}-\y_{t}^\ast\|_2\leq\sqrt{F_{t-1}(\y_{t})-F_{t-1}(\y_{t}^\ast)}\leq\sqrt{\gamma}T^{-\alpha/2}
\end{split}
\end{equation}
where the last inequality is due to \[F_{t-1}(\y_{t})-F_{t-1}(\y_{t}^\ast)\leq \gamma T^{-\alpha}.\]
By combining (\ref{short_eq2}) and (\ref{thm0_eq5}), we have
\begin{equation}
\label{short_eq3}
\begin{split}
\sum_{t=1}^T\left(\|\y_{\tau_{c_t}}-\y_{\tau_{c_t}}^\ast\|_2+\|\y_t^\ast-\y_{t}\|_2\right)\leq \sum_{t=1}^T2\sqrt{\gamma}T^{-\alpha/2}=2\sqrt{\gamma}T^{1-\alpha/2}.
\end{split}
\end{equation}
Next, we proceed to bound the term $\sum_{t=1}^T\|\y_{\tau_{c_t}}^\ast-\y_{t}^\ast\|_2$ in (\ref{short_eq1}). Note that for any $i\geq j\geq1$, we have
\begin{equation}
\label{thm0_eq6}
\begin{split}
\|\y_{j}^\ast-\y_{i}^\ast\|_2^2\leq&F_{i-1}(\y_{j}^\ast)-F_{i-1}(\y_{i}^\ast)\\
=&F_{j-1}(\y_{j}^\ast)-F_{j-1}(\y_{i}^\ast)+\left\langle\eta\sum_{k=j}^{i-1}\nabla f_{c_k}(\x_{c_k}),\y_{j}^\ast-\y_{i}^\ast\right\rangle\\
\leq&\eta\left\|\sum_{k=j}^{i-1}\nabla f_{c_k}(\x_{c_k})\right\|_2\|\y_{j}^\ast-\y_{i}^\ast\|_2\\
\leq&\eta G(i-j)\|\y_{i}^\ast-\y_{j}^\ast\|_2
\end{split}
\end{equation}
where the first inequality is still due to (\ref{cor_scvx}) and the last inequality is due to Assumption \ref{assum1}.

By combining (\ref{short_eq2}) and (\ref{thm0_eq6}), we have
\begin{equation}
\label{thm0_eq7}
\begin{split}
\sum_{t=1}^T\|\y_{\tau_{c_t}}^\ast-\y_{t}^\ast\|_2\leq&\eta G\sum_{t=1}^T\left(t-\tau_{c_t}\right)=\eta G\left(\sum_{t=1}^Tt-\sum_{t=1}^T\tau_{c_t}\right)\\
=&\eta G\left(\sum_{t=1}^Tt-\sum_{t=1}^T\tau_{t}\right)=\eta G\sum_{t=1}^T\left(t-1-\sum_{i=1}^{t-1}|\F_i|\right).
\end{split}
\end{equation}
where the second equality is due to the fact that $c_1,\dots,c_T$ is a permutation of $1,\dots,T$. 

Note that $t-1-\sum_{i=1}^{t-1}|\F_i|$ denotes the number of gradients that have
been queried, but still undelivered at the end of round $t-1$. It is easy to verify that 
\begin{equation}
\label{new_total_delay}
\sum_{t=1}^T\left(t-1-\sum_{i=1}^{t-1}|\F_i|\right)\leq\sum_{t=1}^Td_t=\bar{d}T.
\end{equation}
Finally, by combining (\ref{short_eq1}), (\ref{short_eq3}), (\ref{thm0_eq7}), and (\ref{new_total_delay}), we complete this proof.

\subsection{Proof of Lemma~\ref{thm0_lem1}}
For brevity, we define $h_{t}=F_{t-1}(\y_{t})-F_{t-1}(\y_t^\ast)$ for $t=1,\dots,T+1$ and $h_t(\y_{t-1})=F_{t-1}(\y_{t-1})-F_{t-1}(\y_t^\ast)$ for $t=2,\dots,T+1$. 

For $t=1$, since $\y_1=\argmin_{\y\in\K}\|\y-\y_1\|_2^2$, we have
\begin{equation}
\label{thm0_lem1_eq1}
h_1=F_{0}(\y_1)-F_{0}(\y_1^\ast)=0\leq\frac{4D^2}{\sqrt{T}}.
\end{equation}
Then, for any $T+1\geq t\geq2$, we have
\begin{equation}
\label{thm0_lem1_eq2_0}
\begin{split}
h_t(\y_{t-1})=&F_{t-1}(\y_{t-1})-F_{t-1}(\y_t^\ast)\\
=&F_{t-2}(\y_{t-1})-F_{t-2}(\y_t^\ast)+\langle\eta\nabla f_{c_{t-1}}(\x_{c_{t-1}}),\y_{t-1}-\y_t^\ast\rangle\\
\leq& F_{t-2}(\y_{t-1})-F_{t-2}(\y_{t-1}^\ast)+\langle\eta\nabla f_{c_{t-1}}(\x_{c_{t-1}}),\y_{t-1}-\y_t^\ast\rangle\\
\leq& h_{t-1}+\eta\|\nabla f_{c_{t-1}}(\x_{c_{t-1}})\|_2\|\y_{t-1}-\y_t^\ast\|_2\\
\leq&h_{t-1}+\eta G\|\y_{t-1}-\y_{t-1}^\ast\|_2+\eta G\|\y_{t-1}^\ast-\y_t^\ast\|_2
\end{split}
\end{equation}
where the first inequality is due to $\y_{t-1}^\ast=\argmin_{\y\in\K}F_{t-2}(\y)$, and the last inequality is due to Assumption \ref{assum1} and $\|\y_{t-1}-\y_t^\ast\|_2\leq\|\y_{t-1}-\y_{t-1}^\ast\|_2+\|\y_{t-1}^\ast-\y_t^\ast\|_2$.

Moreover, for any $T+1\geq t\geq2$, we note that $F_{t-2}(\x)$ is $2$-strongly convex, which implies that
\begin{equation}
\label{thm0_lem1_eq2_1}
\|\y_{t-1}-\y_{t-1}^\ast\|_2\leq\sqrt{F_{t-2}(\y_{t-1})-F_{t-2}(\y_{t-1}^\ast)}\leq\sqrt{h_{t-1}}
\end{equation}
where the first inequality is due to (\ref{cor_scvx}).

By using (\ref{cor_scvx}) again, for any $T+1\geq t\geq2$, we also have
\begin{align*}
\|\y_{t-1}^\ast-\y_t^\ast\|_2^2\leq& F_{t-1}(\y_{t-1}^\ast)-F_{t-1}(\y_t^\ast)\\
=&F_{t-2}(\y_{t-1}^\ast)-F_{t-2}(\y_{t}^\ast)+\langle\eta\nabla f_{c_{t-1}}(\x_{c_{t-1}}),\y_{t-1}^\ast-\y_{t}^\ast\rangle\\
\leq&\eta\|\nabla f_{c_{t-1}}(\x_{c_{t-1}})\|_2\|\y_{t-1}^\ast-\y_{t}^\ast\|_2
\end{align*}
which implies that
\begin{equation}
\label{thm0_lem1_eq2_2}
\|\y_{t-1}^\ast-\y_t^\ast\|_2\leq\eta\|\nabla f_{c_{t-1}}(\x_{c_{t-1}})\|_2\leq\eta G
\end{equation}
where the last inequality is due to Assumption \ref{assum1}.

Then, by combining (\ref{thm0_lem1_eq2_0}), (\ref{thm0_lem1_eq2_1}), and (\ref{thm0_lem1_eq2_2}), for any $T+1\geq t\geq2$, we have
\begin{equation}
\label{thm0_lem1_eq2}
h_t(\y_{t-1})\leq h_{t-1}+\eta G\sqrt{h_{t-1}}+\eta^2G^2.
\end{equation}
Then, for any $T+1\geq t\geq2$, we note that $F_{t-1}(\y)$ is also $2$-smooth. It is not hard to verify that
\begin{equation}
\label{thm0_lem1_eq3}
\begin{split}
h_t=&F_{t-1}(\y_{t})-F_{t-1}(\y_t^\ast)\\
=&F_{t-1}(\y_{t-1}+\sigma_{t-1}(\v_{t-1}-\y_{t-1}))-F_{t-1}(\y_t^\ast)\\
\leq&h_t(\y_{t-1})+\langle\nabla F_{t-1}(\y_{t-1}),\sigma_{t-1}(\v_{t-1}-\y_{t-1})\rangle+\sigma_{t-1}^2\|\v_{t-1}-\y_{t-1}\|_2^2.
\end{split}
\end{equation}
Moreover, for any $t\in[T]$, according to Algorithm \ref{alg1}, we have
\begin{equation}
\label{thm0_lem1_eq3_1}
\sigma_{t}=\argmin_{\sigma\in[0,1]}\langle\sigma(\v_{t}-\y_{t}),\nabla F_{t}(\y_{t})\rangle+\sigma^2\|\v_{t}-\y_{t}\|_2^2.
\end{equation}
Then, for any $t=2,\dots,T+1$, by defining $\sigma_{t-1}^\prime=\frac{1}{\sqrt{T}}$ and assuming $h_{t-1}\leq\frac{4D^2}{\sqrt{T}}$, we have
\begin{equation}
\label{thm0_lem1_eq5}
\begin{split}
h_t
\leq&h_t(\y_{t-1})+\langle\nabla F_{t-1}(\y_{t-1}),\sigma_{t-1}^{\prime}(\v_{t-1}-\y_{t-1})\rangle+(\sigma_{t-1}^{\prime})^2\|\v_{t-1}-\y_{t-1}\|_2^2\\
\leq&h_t(\y_{t-1})+\langle\nabla F_{t-1}(\y_{t-1}),\sigma_{t-1}^{\prime}(\y_t^\ast-\y_{t-1})\rangle+(\sigma_{t-1}^{\prime})^2\|\v_{t-1}-\y_{t-1}\|_2^2\\
\leq&(1-\sigma_{t-1}^\prime)h_t(\y_{t-1})+(\sigma_{t-1}^{\prime})^2\|\v_{t-1}-\y_{t-1}\|_2^2\\
\leq&(1-\sigma_{t-1}^\prime)(h_{t-1}+\eta G\sqrt{h_{t-1}}+\eta^2G^2)+(\sigma_{t-1}^{\prime})^2D^2\\
\leq&(1-\sigma_{t-1}^\prime)h_{t-1}+\eta G\sqrt{h_{t-1}}+\eta^2G^2+(\sigma_{t-1}^{\prime})^2D^2\\
\leq&\left(1-\frac{1}{\sqrt{T}}\right)\frac{4D^2}{\sqrt{T}}+\frac{2D^2}{T} +\frac{D^2}{T^{3/2}}+\frac{D^2}{T}\leq\frac{4D^2}{\sqrt{T}}
\end{split}
\end{equation}
where the first inequality is due to (\ref{thm0_lem1_eq3}) and (\ref{thm0_lem1_eq3_1}), the second inequality is due to $\v_{t-1}\in\argmin_{\y\in\K}\langle\nabla F_{t-1}(\y_{t-1}),\y\rangle$, the third inequality is due to the convexity of $F_{t-1}(\y)$, and the fourth inequality is due to (\ref{thm0_lem1_eq2}) and Assumption \ref{assum2}.

By combining (\ref{thm0_lem1_eq1}) and (\ref{thm0_lem1_eq5}), we complete this proof.

\subsection{Proof of Lemma \ref{thm0_lem0}}
In the beginning, we define $\y_{t}^\ast=\argmin_{\y\in \K}F_{t-1}(\y)$ for any $t\in[T+1]$, where $F_t(\y)=\eta\sum_{i=1}^t\langle\nabla f_{c_i}(\x_{c_i}),\y\rangle+\|\y-\y_1\|_2^2$. Then, it is easy to verify that
\begin{equation}
\label{thm0_lem0_eq1}
\begin{split}
\sum_{t=1}^{T}\langle\nabla f_{c_t}(\x_{c_t}),\y_{t}-\x^\ast\rangle=\sum_{t=1}^T\langle\nabla f_{c_t}(\x_{c_t}),\y_t-\y_t^\ast\rangle+\sum_{t=1}^T\langle\nabla f_{c_t}(\x_{c_t}),\y_t^\ast-\x^\ast\rangle.
\end{split}
\end{equation}
Moreover, because of (\ref{cor_scvx}) and Assumption \ref{assum1}, we have
\begin{equation}
\label{thm0_lem0_eq2}
\begin{split}
\sum_{t=1}^T\langle\nabla f_{c_t}(\x_{c_t}),\y_t-\y_t^\ast\rangle\leq&\sum_{t=1}^T\|\nabla f_{c_t}(\x_{c_t})\|_2\|\y_t-\y_t^\ast\|_2\\
\leq&\sum_{t=1}^TG\sqrt{F_{t-1}(\y_t)-F_{t-1}(\y_t^\ast)}.
\end{split}
\end{equation}
Then, to bound $\sum_{t=1}^T\langle\nabla f_{c_t}(\x_{c_t}),\y_t^\ast-\x^\ast\rangle$ in the right side of (\ref{thm0_lem0_eq1}), we introduce the following lemma.
\begin{lem}
\label{lem_garber16}
(Lemma 6.6 of \citet{Garber16})
Let $\{f_t(\y)\}_{t=1}^T$ be a sequence of loss functions and let $\y_t^\ast\in\argmin_{\y\in\K}\sum_{i=1}^tf_{i}(\y)$ for any $t\in[T]$. Then, it holds that
\[\sum_{t=1}^Tf_t(\y_t^\ast)-\min_{\y\in\K}\sum_{t=1}^Tf_t(\y)\leq0.\]
\end{lem}
To apply Lemma~\ref{lem_garber16}, we define \[\tilde{f}_1(\y)=\eta\langle\nabla f_{c_1}(\x_{c_1}),\y\rangle+\|\y-\y_1\|_2^2\] and $\tilde{f}_t(\y)=\eta\langle\nabla f_{c_t}(\x_{c_t}),\y\rangle$ for any $t\geq2$.
We note that $F_{t}(\y)=\sum_{i=1}^t\f_i(\y)$ and $\y_{t+1}^\ast=\argmin_{\y\in \K}F_{t}(\y)$ for any $t=1,\dots,T$. Then, by applying Lemma~\ref{lem_garber16} to $\{\f_t(\y)\}_{t=1}^T$, we have
\begin{equation*}
\begin{split}
\sum_{t=1}^T\f_t(\y_{t+1}^\ast)-\sum_{t=1}^T\f_t(\x^\ast)\leq 0
\end{split}
\end{equation*}
which implies that
\begin{equation*}
\begin{split}
\eta\sum_{t=1}^T\langle\nabla f_{c_t}(\x_{c_t}),\y_{t+1}^\ast-\x^\ast\rangle\leq\|\x^\ast-\y_1\|_2^2-\|\y^\ast_2-\y_1\|_2^2.
\end{split}
\end{equation*}
According to Assumption \ref{assum2}, we have
\begin{equation*}
\begin{split}
\sum_{t=1}^T\langle\nabla f_{c_t}(\x_{c_t}),\y_{t+1}^\ast-\x^\ast\rangle\leq\frac{1}{\eta}\|\x^\ast-\y_1\|_2^2\leq\frac{D^2}{\eta}.
\end{split}
\end{equation*}
Then, we have
\begin{equation}
\label{thm0_lem0_eq4}
\begin{split}
\sum_{t=1}^T\langle\nabla f_{c_t}(\x_{c_t}),\y_t^\ast-\x^\ast\rangle=&\sum_{t=1}^T\langle\nabla f_{c_t}(\x_{c_t}),\y_{t+1}^\ast-\x^\ast\rangle+\sum_{t=1}^T\langle\nabla f_{c_t}(\x_{c_t}),\y_{t}^\ast-\y_{t+1}^\ast\rangle\\
\leq&\frac{D^2}{\eta}+\sum_{t=1}^T\|\nabla f_{c_t}(\x_{c_t})\|_2\|\y_{t}^\ast-\y_{t+1}^\ast\|_2\\
\leq&\frac{D^2}{\eta}+\eta TG^2
\end{split}
\end{equation}
where the second inequality is due to (\ref{thm0_lem1_eq2_2}) and Assumption \ref{assum1}.

By substituting (\ref{thm0_lem0_eq2}) and (\ref{thm0_lem0_eq4}) into (\ref{thm0_lem0_eq1}), we have
\begin{equation}
\label{thm0_lem0_eqfinal}
\begin{split}
\sum_{t=1}^{T}\langle\nabla f_{c_t}(\x_{c_t}),\y_{t}-\x^\ast\rangle\leq&\sum_{t=1}^TG\sqrt{F_{t-1}(\y_t)-F_{t-1}(\y_t^\ast)}+\frac{D^2}{\eta}+\eta TG^2.
\end{split}
\end{equation}
Finally, we can complete this proof by combining (\ref{thm0_lem0_eqfinal}) with Lemma \ref{thm0_lem1} and $\eta=\frac{D}{GT^{3/4}}$.

\subsection{Proof of Theorem \ref{thm1}}
Since $f_t(\x)$ is $\beta$-strongly convex, we have
\begin{equation}
\label{thm1_eq1}
\begin{split}
\sum_{t=1}^Tf_t(\x_t) - \sum_{t=1}^Tf_t(\x^\ast)\leq \sum_{t=1}^T\langle\nabla f_t(\x_t),\x_t-\x^\ast\rangle-\sum_{t=1}^T\frac{\beta}{2}\|\x_t-\x^\ast\|_2^2.
\end{split}
\end{equation}
Then, we note that the first term in the right side of (\ref{thm1_eq1}) can be bounded by reusing (\ref{thm0_eq1}). Specifically, we have
\begin{equation}
\label{thm1_eq2}
\begin{split}
\sum_{t=1}^Tf_t(\x_t) - \sum_{t=1}^Tf_t(\x^\ast)
\leq&\sum_{t=1}^T\langle\nabla f_{c_t}(\x_{c_t}),\y_{t}-\x^\ast\rangle+\sum_{t=1}^TG\|\y_{\tau_{c_t}}-\y_{t}\|_2-\sum_{t=1}^T\frac{\beta}{2}\|\x_t-\x^\ast\|_2^2\\
=&\sum_{t=1}^T\langle\nabla f_{c_t}(\x_{c_t}),\y_{t}-\x^\ast\rangle+\sum_{t=1}^TG\|\y_{\tau_{c_t}}-\y_{t}\|_2-\sum_{t=1}^T\frac{\beta}{2}\|\x_{c_t}-\x^\ast\|_2^2
\end{split}
\end{equation}
where the equality is due to the fact that $c_1,\dots,c_T$ is a permutation of $1,\dots,T$.

Next, we consider the last term in the right side of (\ref{thm1_eq2}). 
For any $\y_t,\x_{c_t},\x^\ast\in\K$, we have
\begin{equation*}
\begin{split}
\|\y_t-\x^\ast\|_2^2=&\|\y_t-\x_{c_t}\|_2^2+\|\x_{c_t}-\x^\ast\|_2^2+2\langle\y_t-\x_{c_t},\x_{c_t}-\x^\ast\rangle\\
\leq&3D\|\y_t-\x_{c_t}\|_2+\|\x_{c_t}-\x^\ast\|_2^2
\end{split}
\end{equation*}
where the last inequality is due to \[2\langle\y_t-\x_{c_t},\x_{c_t}-\x^\ast\rangle\leq2\|\y_t-\x_{c_t}\|_2\|\x_{c_t}-\x^\ast\|_2\] and Assumption \ref{assum2}.

By combining the above inequality and (\ref{re_index0}) with (\ref{thm1_eq2}), we have
\begin{equation}
\label{thm1_eq4}
\begin{split}
&\sum_{t=1}^Tf_t(\x_t) - \sum_{t=1}^Tf_t(\x^\ast)\\
\leq&\sum_{t=1}^T\langle\nabla f_{c_t}(\x_{c_t}),\y_{t}-\x^\ast\rangle-\sum_{t=1}^T\frac{\beta}{2}\|\y_{t}-\x^\ast\|_2^2+\sum_{t=1}^T\left(G+\frac{3\beta D}{2}\right)\|\y_{\tau_{c_t}}-\y_{t}\|_2.
\end{split}
\end{equation}
Then, we first consider the last term in (\ref{thm1_eq4}), and establish the following lemma by carefully analyzing the distance $\|\y_{\tau_{c_t}}-\y_{t}\|_2$.
\begin{lem}
\label{lem_sc_sum_terms}
Let $\y_{t}^\ast=\argmin_{\y\in \K}F_{t-1}(\y)$ for any $t=2,\dots,T+1$, where $F_{t}(\y)$ is defined in (\ref{sec4-1-eq2}). Suppose Assumptions \ref{assum1} and \ref{assum2} hold, all losses are $\beta$-strongly convex, and there exist some constants $\gamma>0$ and $0\leq\alpha<1$ such that $F_{t-1}(\y_{t})-F_{t-1}(\y_t^\ast)\leq \gamma(t-1)^{\alpha}$ for any $t=2,\dots,T+1$. Algorithm \ref{alg2} ensures
\begin{equation*}
\begin{split}
\sum_{t=1}^T\|\y_{\tau_{c_t}}-\y_{t}\|_2\leq&6{d}D+2{d}\sqrt{\frac{2\gamma}{\beta}}+\sqrt{\frac{2\gamma}{\beta}}\frac{4}{\alpha+1}T^{(1+\alpha)/2}+\frac{4{d}(G+\beta D)}{\beta}\ln T.
\end{split}
\end{equation*}
\end{lem}
Note that Lemma \ref{lem_sc_sum_terms} also introduces an assumption about $\y_t$ and $F_{t-1}(\y)$. According to our Algorithm \ref{alg2}, $\y_t$ is actually generated by approximately minimizing $F_{t-1}(\y)$ with a linear optimization step. Therefore, by following the analysis of OFW for strongly convex losses \citep{Wan-AAAI-2021-C}, we show that this assumption is satisfied with $\gamma=16(G+2\beta D)^2/\beta$ and $\alpha=1/3$.
\begin{lem}
\label{thm1_SC_OFW1}
Let $\y_{t}^\ast=\argmin_{\y\in \K}F_{t-1}(\y)$ for any $t=2,\dots,T+1$, where $F_{t}(\y)$ is defined in (\ref{sec4-1-eq2}). Suppose Assumptions \ref{assum1} and \ref{assum2} hold, and all losses are $\beta$-strongly convex. For any $t=2,\dots,T+1$, Algorithm~\ref{alg2} has
\[F_{t-1}(\y_{t})-F_{t-1}(\y_t^\ast)\leq\frac{16(G+2\beta D)^2(t-1)^{1/3}}{\beta}.\]
\end{lem}
By combining Lemmas \ref{lem_sc_sum_terms} and \ref{thm1_SC_OFW1}, we have
\begin{equation}
\label{final_eq1}
\begin{split}
&\sum_{t=1}^T\left(G+\frac{3\beta D}{2}\right)\|\y_{\tau_{c_t}}-\y_{t}\|_2\\
\leq&(C_1+C_2)\left(3{d}D+\frac{4\sqrt{2}{d}C_2}{\beta}+\frac{6\sqrt{2}C_2T^{2/3}}{\beta}+\frac{2{d}C_1}{\beta}\ln T\right)
\end{split}
\end{equation}
where $C_1=G+\beta D$ and $C_2=G+2\beta D$.

Moreover, by following the analysis of OFW for strongly convex losses \citep{Wan-AAAI-2021-C}, we can also establish the following lemma.
\begin{lem}
\label{thm1_SC_OFW2}
Suppose Assumptions \ref{assum1} and \ref{assum2} hold, and all losses are $\beta$-strongly convex. For any $\x^\ast\in\K$, Algorithm~\ref{alg2} ensures
\begin{equation*}
\begin{split}
&\sum_{t=1}^T\langle\nabla f_{c_t}(\x_{c_t}),\y_{t}-\x^\ast\rangle-\sum_{t=1}^T\frac{\beta}{2}\|\y_{t}-\x^\ast\|_2^2\\
\leq& \frac{6\sqrt{2}(G+2\beta D)^2T^{2/3}}{\beta}+\frac{2(G+2\beta D)^2\ln T}{\beta}+(G+\beta D)D.
\end{split}
\end{equation*}
\end{lem}
Finally, by combining (\ref{thm1_eq4}), (\ref{final_eq1}), and Lemma \ref{thm1_SC_OFW2}, we have
\begin{equation*}
\begin{split}
\sum_{t=1}^Tf_t(\x_t)-\sum_{t=1}^Tf_t(\x^\ast)\leq&\frac{6\sqrt{2}C_2(C_1+2C_2)T^{2/3}}{\beta}+\frac{2C_2^2\ln T}{\beta}+C_1D\\
&+(C_1+C_2)\left(3{d}D+\frac{4\sqrt{2}{d}C_2}{\beta}+\frac{2{d}C_1}{\beta}\ln T\right)
\end{split}
\end{equation*}
which completes the proof.

\subsection{Proof of Lemma \ref{lem_sc_sum_terms}}
If $T\leq 2{d}$, according to Assumption \ref{assum2}, it is easy to verify that
\begin{equation}
\label{thm1_short_eq0}
\sum_{t=1}^T\|\y_{\tau_{c_t}}-\y_{t}\|_2\leq2{d}D.
\end{equation}
Therefore, in the following, we only need to consider the case with $T>2{d}$.

Let $\y_1^\ast=\y_1$. It is easy to verify 
\begin{equation}
\label{thm1_short_eq1}
\begin{split}
\sum_{t=1}^T\|\y_{\tau_{c_t}}-\y_{t}\|_2\leq&\sum_{t=1}^T\left(\|\y_{\tau_{c_t}}-\y_{\tau_{c_t}}^\ast\|_2+\|\y_{\tau_{c_t}}^\ast-\y_{t}^\ast\|_2+\|\y_t^\ast-\y_{t}\|_2\right)\\
=&\sum_{t=1}^T\left(\|\y_{\tau_{t}}-\y_{\tau_{t}}^\ast\|_2+\|\y_{\tau_{c_t}}^\ast-\y_{t}^\ast\|_2+\|\y_t^\ast-\y_{t}\|_2\right)\\
\leq&6{d}D+\sum_{t=2{d}+1}^{T}\left(\|\y_{\tau_{t}}-\y_{\tau_{t}}^\ast\|_2+\|\y_{\tau_{c_t}}^\ast-\y_{t}^\ast\|_2+\|\y_t^\ast-\y_{t}\|_2\right)
\end{split}
\end{equation}
where the equality is due to the fact that $c_1,\dots,c_T$ is a permutation of $1,\dots,T$, and the last inequality is due to Assumption \ref{assum2}.


According to (\ref{sec4-1-eq2}), it easy to verify that $F_{t-1}(\y)$ is $(t-1)\beta$-strongly convex for any $t=2,\dots,T+1$. Therefore, for any $t=2,\dots,T+1$, we have
\begin{equation}
\label{thm1_eq7}
\begin{split}
\|\y_t-\y_t^\ast\|_2\leq\sqrt{\frac{2(F_{t-1}(\y_t)-F_{t-1}(\y_t^\ast))}{(t-1)\beta}}\leq \sqrt{\frac{2\gamma}{(t-1)^{1-\alpha}\beta}}
\end{split}
\end{equation}
where the first inequality is due to (\ref{cor_scvx}) and the second inequality is due to \[F_{t-1}(\y_{t})-F_{t-1}(\y_t^\ast)\leq \gamma(t-1)^{\alpha}.\]
Then, we have
\begin{equation}
\label{thm1_short_eq3-pre}
\begin{split}
\sum_{t=2{d}+1}^{T}\left(\|\y_{\tau_{t}}-\y_{\tau_{t}}^\ast\|_2+\|\y_t^\ast-\y_{t}\|_2\right)\leq&\sum_{t=2{d}+1}^{T}\left(\|\y_{\tau_{t}}-\y_{\tau_{t}}^\ast\|_2+\|\y_t^\ast-\y_{t}\|_2\right)\\
\leq&\sum_{t=2{d}+1}^{T}\left(\sqrt{\frac{2\gamma}{(\tau_t-1)^{1-\alpha}\beta}}+\sqrt{\frac{2\gamma}{(t-1)^{1-\alpha}\beta}}\right)\\
\leq&\sum_{t=2{d}+1}^{T}2\sqrt{\frac{2\gamma}{(\tau_t-1)^{1-\alpha}\beta}}
\end{split}
\end{equation}
where the second inequality is due to (\ref{thm1_eq7}) and $\tau_t>1$, and the third inequality is due to $\tau_t\leq t$ and $\alpha<1$. 

To further bound the right side of (\ref{thm1_short_eq3-pre}), we introduce the following lemma.
\begin{lem}
\label{thm0_lem3}
Let $\tau_t=1+\sum_{i=1}^{t-1}|\F_i|$ for any $t\in[T+{d}-1]$. If $T>2d$, for $0<\alpha\leq1$, we have
\begin{equation*}
\sum_{t=2{d}+1}^T(\tau_t-1)^{-\alpha/2}\leq {d}+\frac{2}{2-\alpha}T^{1-\alpha/2}.
\end{equation*}
\end{lem}
By combining Lemma \ref{thm0_lem3} with (\ref{thm1_short_eq3-pre}), we have
\begin{equation}
\label{thm1_short_eq3}
\begin{split}
\sum_{t=2{d}+1}^{T}\left(\|\y_{\tau_{t}}-\y_{\tau_{t}}^\ast\|_2+\|\y_t^\ast-\y_{t}\|_2\right)\leq2{d}\sqrt{\frac{2\gamma}{\beta}}+\sqrt{\frac{2\gamma}{\beta}}\frac{4}{\alpha+1}T^{(1+\alpha)/2}.
\end{split}
\end{equation}
Before considering $\|\y_{\tau_{c_t}}^\ast-\y_{t}^\ast\|_2$, we define \[\tilde{f}_t(\y)=\langle \nabla f_{c_t}(\x_{c_t}),\y\rangle+\frac{\beta}{2}\|\y-\y_t\|_2^2\] for any $t\in[T]$. Then, for any $\x,\y\in\K$ and $t\in[T]$, we have
\begin{equation}
\label{thm1_eq5_6}
\begin{split}
|\tilde{f}_t(\x)-\tilde{f}_t(\y)|=&\left|\langle \nabla f_{c_t}(\x_{c_t}),\x-\y\rangle+\frac{\beta}{2}\|\x-\y_t\|_2^2-\frac{\beta}{2}\|\y-\y_t\|_2^2\right|\\
=&\left|\langle \nabla f_{c_t}(\x_{c_t}),\x-\y\rangle+\frac{\beta}{2}\langle\x-\y_t+\y-\y_t,\x-\y\rangle\right|\\
\leq&\|\nabla f_{c_t}(\x_{c_t})\|_2\|\x-\y\|_2+\frac{\beta}{2}(\|\x-\y_t\|_2+\|\y-\y_t\|_2)\|\x-\y\|_2\\
\leq&(G+\beta D)\|\x-\y\|_2
\end{split}
\end{equation}
where the last inequality is due to Assumptions \ref{assum1} and \ref{assum2}.

Note that $F_t(\y)=\sum_{i=1}^t\tilde{f}_i(\y)$. Because of (\ref{cor_scvx}), for any $i\geq j>1$, we have
\begin{equation}
\label{thm1_eq7_8}
\begin{split}
\|\y_{j}^\ast-\y_{i}^\ast\|_2^2\leq&\frac{2(F_{i-1}(\y_{j}^\ast)-F_{i-1}(\y_{i}^\ast))}{(i-1)\beta}\\
=&\frac{2(F_{j-1}(\y_{j}^\ast)-F_{j-1}(\y_{i}^\ast))+2\sum_{k=j}^{i-1}\left(\tilde{f}_k(\y_{j}^\ast)-\tilde{f}_k(\y_{i}^\ast)\right)}{(i-1)\beta}\\
\leq&\frac{2(i-j)(G+\beta D)\|\y_{j}^\ast-\y_{i}^\ast\|_2}{(i-1)\beta}
\end{split}
\end{equation}
where the last inequality is due to $\y_j^\ast=\argmin_{\y\in\K}F_{j-1}(\y)$ and (\ref{thm1_eq5_6}).

Moreover, we introduce two useful properties of $\tau_{c_t}$. First, since the gradient $\nabla f_{c_t}(\x_{c_t})$ arrives at the end of round $c_t+d_{c_t}-1$ and is the $t$-th used gradient, we have
\begin{equation}
\label{short_eq2-x}
t\leq c_t+d_{c_t}-1\leq c_t+{d}-1.
\end{equation}
Second, by combing (\ref{short_eq2-x}) with the fact that all gradients queried at rounds $1,\dots,t-2{d}$ must arrive before round $t-{d}$, for any $t\geq2{d}+1$, we have \begin{equation}
\label{short_eq2-3}
\tau_{c_t}=1+\sum_{k=1}^{c_t-1}|\F_k|\geq1+\sum_{k=1}^{t-{d}}|\F_k|\geq t-2{d}+1>t-2{d}.
\end{equation}
Then, it is not hard to verify that
\begin{equation}
\label{thm1_eq8}
\begin{split}
\sum_{t=2{d}+1}^T\|\y_{\tau_{c_t}}^\ast-\y_{t}^\ast\|_2\leq&\sum_{t=2{d}+1}^T\frac{2(t-\tau_{c_t})(G+\beta D)}{(t-1)\beta}\\
\leq&\sum_{t=2{d}+1}^T\frac{4{d}(G+\beta D)}{(t-1)\beta}\\
\leq&\frac{4{d}(G+\beta D)}{\beta}\sum_{t=2}^T\frac{1}{t}\\
\leq&\frac{4{d}(G+\beta D)}{\beta}\ln T.
\end{split}
\end{equation}
where the first inequality is due to (\ref{short_eq2}), (\ref{short_eq2-3}), and (\ref{thm1_eq7_8}), and the second inequality is due to (\ref{short_eq2-3}).

By combining (\ref{thm1_short_eq1}) with (\ref{thm1_short_eq3}) and (\ref{thm1_eq8}), if $T>2{d}$, we have
\begin{equation}
\label{thm1_eq10}
\begin{split}
\sum_{t=1}^T\|\y_{\tau_{c_t}}-\y_{t}\|_2\leq6{d}D+2{d}\sqrt{\frac{2\gamma}{\beta}}+\sqrt{\frac{2\gamma}{\beta}}\frac{4}{\alpha+1}T^{(1+\alpha)/2}+\frac{4{d}(G+\beta D)}{\beta}\ln T.
\end{split}
\end{equation}
Finally, we complete the proof by combining (\ref{thm1_eq10}) with (\ref{thm1_short_eq0}).

\subsection{Proof of Lemmas \ref{thm1_SC_OFW1} and \ref{thm1_SC_OFW2}}
Let $
\tilde{f}_t(\y)=\langle \nabla f_{c_t}(\x_{c_t}),\y\rangle+\frac{\beta}{2}\|\y-\y_t\|_2^2$
for any $t=1,\dots,T$, which is $\beta$-strongly convex. As proved in (\ref{thm1_eq5_6}), functions $\tilde{f}_1(\y),\dots,\tilde{f}_T(\y)$ are $(G+\beta D)$-Lipschitz over $\K$ (see the definition of Lipschitz functions in \citet{Hazan2016}). Then, recall that $F_\tau(\y)$ defined in Algorithm \ref{alg2} is equivalent to that defined in (\ref{sec4-1-eq2}), and thus we have $F_\tau(\y)=\sum_{i=1}^\tau\tilde{f}_i(\y)$. Moreover, because of $\nabla\tilde{f}_t(\y_t)=\nabla f_{c_t}(\x_{c_t})$, it is not hard to verify that decisions $\y_1,\dots,\y_{T+1}$ in our Algorithm \ref{alg2} are actually generated by performing OFW for strongly convex losses (see Algorithm 2 in \citet{Wan-AAAI-2021-C} for details) on functions $\tilde{f}_1(\y),\dots,\tilde{f}_T(\y)$. As a result, we can derive Lemmas \ref{thm1_SC_OFW1} and \ref{thm1_SC_OFW2} from existing theoretical guarantees of OFW for strongly convex losses.

Specifically, by applying Lemma 6 of \citet{Wan-AAAI-2021-C} with functions $\tilde{f}_1(\y),\dots,\tilde{f}_T(\y)$, we have
\[
F_{t-1}(\y_{t})-F_{t-1}(\y_t^\ast)\leq\frac{16(G+2\beta D)^2(t-1)^{1/3}}{\beta}
\]
for any $t=2,\dots,T+1$, which completes the proof of our Lemma \ref{thm1_SC_OFW1}.

Then, by applying Theorem 3 of \citet{Wan-AAAI-2021-C} with functions $\tilde{f}_1(\y),\dots,\tilde{f}_T(\y)$, we have
\begin{equation}
\label{Wan-eq}
\begin{split}
\sum_{t=1}^{T}\tilde{f}_t(\y_t)-\sum_{t=1}^{T}\tilde{f}_t(\x^\ast)\leq& \frac{6\sqrt{2}(G+2\beta D)^2T^{2/3}}{\beta}+\frac{2(G+2\beta D)^2\ln T}{\beta}+(G+\beta D)D.
\end{split}
\end{equation}
Finally, we complete the proof of Lemma \ref{thm1_SC_OFW2} by combining (\ref{Wan-eq}) with the following equality
\begin{equation}
\label{trans}
\sum_{t=1}^{T}\left(\langle\nabla f_{c_t}(\x_{c_t}),\y_{t}-\x^\ast\rangle-\frac{\beta}{2}\|\y_t-\x^\ast\|_2^2\right)=\sum_{t=1}^{T}\tilde{f}_t(\y_t)-\sum_{t=1}^{T}\tilde{f}_t(\x^\ast).
\end{equation}

\subsection{Proof of Lemma \ref{thm0_lem3}} First, since the gradient $\nabla f_1(\x_1)$ must arrive before round ${d}+1$, for any $T\geq t\geq2{d}+1$, it is easy to verify that 
\begin{equation}
\label{thm1_short_eq2}
\tau_t=1+\sum_{i=1}^{t-1}|\F_i|\geq1+\sum_{i=1}^{{d}+1}|\F_i|\geq2.
\end{equation}
Moreover, for any $i\geq2$ and $(i+1){d}\geq t\geq i{d}+1$, since all gradients queried at rounds $1,\dots,(i-1){d}+1$ must arrive before round $i{d}+1$, we have
\begin{equation}
\label{eq4_123}
\tau_t=1+\sum_{i=1}^{t-1}|\F_i|\geq(i-1){d}+2.
\end{equation}
Then, we have
\begin{align*}
&\sum_{t=2{d}+1}^T(\tau_t-1)^{-\alpha/2}\\
=&\sum_{t=2{d}+1}^{\lfloor T/{d}\rfloor {d}}(\tau_t-1)^{-\alpha/2}+\sum_{t=\lfloor T/{d}\rfloor {d}+1}^{T}(\tau_t-1)^{-\alpha/2}\leq\sum_{i=2}^{\lfloor T/{d}\rfloor-1}\sum_{t=i{d}+1}^{(i+1){d}}(\tau_t-1)^{-\alpha/2}+{d}\\
\leq& {d}+\sum_{i=2}^{\lfloor T/{d}\rfloor-1}{d}((i-1){d}+1)^{-\alpha/2}\leq{d}+\sum_{i=2}^{\lfloor T/{d}\rfloor-1}{d}^{1-\alpha/2}(i-1)^{-\alpha/2}\\
\leq& {d}+\sum_{i=1}^{\lfloor T/{d}\rfloor}{d}^{1-\alpha/2}i^{-\alpha/2}\leq{d}+\frac{2}{2-\alpha}{d}^{1-\alpha/2}\left(\lfloor T/{d}\rfloor\right)^{1-\alpha/2}\\
\leq&{d}+\frac{2}{2-\alpha}T^{1-\alpha/2}
\end{align*}
where the first inequality is due to $(\tau_t-1)^{-\alpha/2}\leq1$ for $\alpha>0$ and $\tau_t\geq2$, and the second inequality is due to (\ref{eq4_123}) and $\alpha>0$.

\subsection{Proof of Theorem \ref{thm1_ospf}}
We start this proof by defining
\begin{equation}
\label{eq1-ospf}
\begin{split}
&\x_m^\ast=\E_{\v\in\B}\left(\argmax_{\x\in\K}\left\langle-\sum_{i=1}^{(m-1)K}\nabla f_i(\x_{m_i})+\frac{\v}{\delta},\x\right\rangle\right),\\
&\x_m^\prime=\E_{\v\in\B}\left(\argmax_{\x\in\K}\left\langle-\hat{\g}_{(m-1)K}+\frac{\v}{\delta},\x\right\rangle\right)
\end{split}
\end{equation}
for each block $m$, where $m_i=\lceil i/K\rceil$ denotes the block index of round $i$. Moreover, let $\xi_m$ denote the randomness introduced at each block $m$ of Algorithm \ref{alg3}, and $\xi_{1:m}$ denote the randomness introduced at the first $m$ blocks.

Then, according to the convexity of loss functions, we have
\begin{equation}
\begin{split}
\label{eq2-ospf}
&\sum_{m=1}^{T/K}\sum_{t=(m-1)K+1}^{mK}\left(f_t(\x_m)-f_t(\x^\ast)\right)\\
\leq&\sum_{m=1}^{T/K}\sum_{t=(m-1)K+1}^{mK}\langle\nabla f_t(\x_m),\x_m-\x^\ast\rangle\\
=&\sum_{m=1}^{T/K}\sum_{t=(m-1)K+1}^{mK}\langle\nabla f_t(\x_m),\x_m-\x_m^\prime\rangle+\sum_{m=1}^{T/K}\sum_{t=(m-1)K+1}^{mK}\langle\nabla f_t(\x_m),\x_m^\prime-\x_m^\ast\rangle\\
&+\sum_{m=1}^{T/K}\sum_{t=(m-1)K+1}^{mK}\langle\nabla f_t(\x_m),\x_m^\ast-\x^\ast\rangle.
\end{split}
\end{equation}
In the following, we provide upper bounds for the three terms in the right side of (\ref{eq2-ospf}) by introducing the following lemmas.
\begin{lem}
\label{lem15_hazan}
(Lemma 15 in \citet{Hazan20}) Let $Z_1,\dots,Z_K$ be i.i.d. samples of a bounded random vector $Z\in\mathbb{R}^n$ with $\|Z\|\leq D$ and $\bar{Z}=\E(Z)$. Let $\bar{Z}_K=\frac{1}{K}\sum_{u=1}^KZ_u$. Then, we have
\[\E_{Z_1,\dots,Z_K}\left(\|\bar{Z}_K-\bar{Z}\|_2^2\right)\leq\frac{4D^2}{K}.\]
\end{lem}
\begin{lem}
\label{lem1_ospf}
Let \[L(\u)=\E_{\v\in\B}\left(\argmax_{\x\in\K}\left\langle\u+\frac{\v}{\delta},\x\right\rangle\right)\] where $\B=\left\{\x\in\mathbb{R}^n|\|\x\|_2\leq1\right\}$. Under Assumption \ref{assum2}, for any $\u$ and $\x$, we have
\[\|L(\u)-L(\x)\|_2\leq\delta nD\|\u-\x\|_2.\]
\end{lem}
\begin{lem}
\label{lem2_ospf}
Let $\ell_1(\x)=\langle\nabla_1,\x\rangle,\dots,\ell_T(\x)=\langle\nabla_T,\x\rangle$ be a sequence of linear loss functions. Define \[\x_t^\ast=\E_{\v\in\B}\left(\argmax_{\x\in\K}\left\langle-\sum_{i=1}^{t-1}\nabla_i+\frac{\v}{\delta},\x\right\rangle\right)\] where $\B=\left\{\x\in\mathbb{R}^n|\|\x\|_2\leq1\right\}$ and $\delta>0$ is a parameter. Suppose Assumption \ref{assum2} holds and $\|\nabla_t\|\leq G$ for any $t=1,\dots,T$. For any $\x^\ast\in\K$, we have
\[
\sum_{t=1}^T\ell_t(\x_{t}^\ast)-\sum_{t=1}^T\ell_t(\x^\ast)\leq \frac{2D}{\delta}+\frac{\delta nDTG^2}{2}.
\]

\end{lem}
Then, for any $m=1,\dots,T/K$, we first have
\begin{equation}
\begin{split}
\label{eq3-ospf}
&\sum_{t=(m-1)K+1}^{mK}\langle\nabla f_t(\x_m),\x_m-\x_m^\prime\rangle\\
=&\sum_{t=(m-1)K+1}^{mK}\langle\nabla f_{t}(\x_m^\prime)+\nabla f_{t}(\x_m)-\nabla f_{t}(\x_m^\prime),\x_m-\x_m^\prime\rangle\\
\leq&\sum_{t=(m-1)K+1}^{mK}(\langle\nabla f_{t}(\x_m^\prime),\x_m-\x_m^\prime\rangle+\|\nabla f_{t}(\x_m)-\nabla f_{t}(\x_m^\prime)\|_2\|\x_m-\x_m^\prime\|_2)\\
\leq&\sum_{t=(m-1)K+1}^{mK}(\langle\nabla f_{t}(\x_m^\prime),\x_m-\x_m^\prime\rangle+\alpha\|\x_m-\x_m^\prime\|_2^2)
\end{split}
\end{equation}
where the last inequality is due to the smoothness of $f_t(\x)$.

According to (\ref{eq1-ospf}) and the assumption that the adversary is oblivious, it is easy to verify that conditioned on the previous randomness $\xi_{1:m-1}$, $\x_m^\prime$ is a deterministic quantity, and $\x_m^j$ is an unbiased estimation of $\x_m^\prime$ for any $j=1,\dots,K$. Therefor, we have
\begin{equation}
\begin{split}
\label{eq4-ospf}
&\E\left(\sum_{t=(m-1)K+1}^{mK}\langle\nabla f_t(\x_m),\x_m-\x_m^\prime\rangle\right)\\
\leq&\E\left(\sum_{t=(m-1)K+1}^{mK}\langle\nabla f_{t}(\x_m^\prime),\x_m-\x_m^\prime\rangle\right)+\E\left(K\alpha\|\x_m-\x_m^\prime\|_2^2\right)\\
=&\E\left(\E\left(\left.\sum_{t=(m-1)K+1}^{mK}\left\langle\nabla f_{t}(\x_m^\prime),\frac{1}{K}\sum_{j=1}^K\x_m^j-\x_m^\prime\right\rangle\right|\xi_{1:m-1}\right)\right)\\
&+\E\left(\E\left(\left.K\alpha\left\|\frac{1}{K}\sum_{j=1}^K\x_m^j-\x_m^\prime\right\|_2^2\right|\xi_{1:m-1}\right)\right)\\
=&\E\left(\E\left(\left.K\alpha\left\|\frac{1}{K}\sum_{j=1}^K\x_m^j-\x_m^\prime\right\|_2^2\right|\xi_{1:m-1}\right)\right)\\
\leq&4\alpha D^2
\end{split}
\end{equation}
where the second inequality is due to the unbiasedness of $\x_m^j$, and the last inequality is due to  Assumption \ref{assum2} and Lemma \ref{lem15_hazan}.

By summing the above inequality over $m=1,\dots,T/K$, we have
\begin{equation}
\begin{split}
\label{eq5-ospf}
\E\left(\sum_{m=1}^{T/K}\sum_{t=(m-1)K+1}^{mK}\langle\nabla f_t(\x_m),\x_m-\x_m^\prime\rangle\right)\leq\frac{4\alpha TD^2}{K}.
\end{split}
\end{equation}
Next, for the second term in the right side of (\ref{eq2-ospf}), we have 
\begin{equation}
\begin{split}
\label{eq6-ospf-pre}
\sum_{m=1}^{T/K}\sum_{t=(m-1)K+1}^{mK}\langle\nabla f_t(\x_m),\x_m^\prime-\x_m^\ast\rangle\leq&\sum_{m=1}^{T/K}\sum_{t=(m-1)K+1}^{mK}\|\nabla f_t(\x_m)\|_2\|\x_m^\prime-\x_m^\ast\|_2\\
\leq&\sum_{m=1}^{T/K}GK\|\x_m^\prime-\x_m^\ast\|_2\\
\leq&\sum_{m=1}^{T/K}\delta nGDK\left\|\hat{\g}_{(m-1)K}-\sum_{i=1}^{(m-1)K}\nabla f_i(\x_{m_i})\right\|_2
\end{split}
\end{equation}
where the second inequality is due to Assumption \ref{assum1}, and the last inequality is due to (\ref{eq1-ospf}) and Lemma \ref{lem1_ospf}.

For brevity, for each block $m$, we further define \[\U_{m}=\{1,\dots,(m-1)K\}\setminus\cup_{t=1}^{(m-1)K}\F_{t}.\]
Then, we can simplified (\ref{eq6-ospf-pre}) as
\begin{equation}
\begin{split}
\label{eq6-ospf}
\sum_{m=1}^{T/K}\sum_{t=(m-1)K+1}^{mK}\langle\nabla f_t(\x_m),\x_m^\prime-\x_m^\ast\rangle
\leq&\sum_{m=1}^{T/K}\delta nGDK\left\|\hat{\g}_{(m-1)K}-\sum_{i=1}^{(m-1)K}\nabla f_i(\x_{m_i})\right\|_2\\
=&\sum_{m=1}^{T/K}\delta nGDK\left\|\sum_{i\in\U_{m}}\nabla f_i(\x_{m_i})\right\|_2\\
\leq&\sum_{m=1}^{T/K}\delta nDKG^2|\U_{m}|.
\end{split}
\end{equation}
We notice that the set $\U_{m}$ actually contains the time stamp of gradients that are queried, but still not
arrive at the end of round $(m-1)K$. Since all gradients queried before round $(m-1)K-{d}+1$ must be received before the end of round $(m-1)K$, it is easy to verify that 
\begin{equation}
\label{eq7-ospf}
|\U_m|\leq {d}.
\end{equation}
By combining (\ref{eq6-ospf}) with (\ref{eq7-ospf}), we have
\begin{equation}
\begin{split}
\label{eq8-ospf}
\sum_{m=1}^{T/K}\sum_{t=(m-1)K+1}^{mK}\langle\nabla f_t(\x_m),\x_m^\prime-\x_m^\ast\rangle\leq\sum_{m=1}^{T/K}\delta n{d}DKG^2=\delta n{d}DTG^2.
\end{split}
\end{equation}
To bound the last term in the right side of (\ref{eq2-ospf}), we define $\nabla_m=\sum_{t=(m-1)K+1}^{mK}\nabla f_t(\x_m)$ and $\ell_m(\x)=\left\langle\nabla_m,\x\right\rangle$ for any $m=1,\dots,T/K$. According to Assumption \ref{assum1}, we have
\[
\|\nabla_m\|_2\leq KG.
\]
Moreover, according to (\ref{eq1-ospf}), we have
\[
\x_m^\ast=\E_{\v\in\B}\left(\argmax_{\x\in\K}\left\langle-\sum_{i=1}^{m-1}\nabla_i+\frac{\v}{\delta},\x\right\rangle\right)
\]
By applying Lemma \ref{lem2_ospf} with loss functions $\ell(\x)_1,\dots,\ell(\x)_{T/K}$, we have
\begin{equation}
\begin{split}
\label{eq9-ospf}
\sum_{m=1}^{T/K}\sum_{t=(m-1)K+1}^{mK}\langle\nabla f_t(\x_m),\x_m^\ast-\x^\ast\rangle=\sum_{m=1}^{T/K}\ell_m(\x_m^\ast)-\sum_{m=1}^{T/K}\ell_m(\x^\ast)\leq\frac{2D}{\delta}+\frac{\delta nDTKG^2}{2}.
\end{split}
\end{equation}
Finally, by substituting (\ref{eq5-ospf}), (\ref{eq8-ospf}), and (\ref{eq9-ospf}) into (\ref{eq2-ospf}), we have
\begin{align*}
\E\left(\sum_{m=1}^{T/K}\sum_{t=(m-1)K+1}^{mK}\left(f_t(\x_m)-f_t(\x^\ast)\right)\right)\leq& \frac{4\alpha TD^2}{K}+\delta n{d}DTG^2+\frac{2D}{\delta}+\frac{\delta nDTKG^2}{2}\\
=&4\alpha D^2T^{2/3}+2\sqrt{n}{d}DGT^{1/3}+2\sqrt{n}DGT^{2/3}
\end{align*}
where the equality is due to $\delta=\frac{2}{\sqrt{n}GT^{2/3}}$ and $K=T^{1/3}$.


\subsection{Proof of Lemma \ref{lem1_ospf}}
We first introduce two lemmas from \citet{Hazan20}.
\begin{lem}
(Lemma 6 in \citet{Hazan20}) Let $\mathcal{M}_{\K}(\y)=\max_{\x\in\K}\langle\y,\x\rangle$. Then, $\mathcal{M}_{\K}(\y)$ is convex, and under Assumption \ref{assum2}, it is $D$-Lipschitz, i.e., for any $\y_1,\y_2$,
\[|\mathcal{M}_{\K}(\y_1)-\mathcal{M}_{\K}(\y_2)|\leq D\|\y_1-\y_2\|_2.\]

\end{lem}
\begin{lem}
(Lemma 11 in \citet{Hazan20}) 
Given a $L$-Lipschitz function $g(\y):\mathbb{R}^n\mapsto\mathbb{R}$, the function \[\hat{g}(\y)=\E_{\v\in\B}\left(g\left(\y+\frac{\v}{\delta}\right)\right)\] is $\delta nL$-smooth.
\end{lem}
By combining these two lemmas, the function $\E_{\v\in\B}\left(\max_{\x\in\K}\left\langle\y+\frac{\v}{\delta},\x\right\rangle\right)$ is $\delta nD$-smooth. Moreover, for $\mathcal{M}_{\K}(\y)=\max_{\x\in\K}\langle\y,\x\rangle$, we notice that \[\nabla\mathcal{M}_{\K}(\y)=\argmax_{\x\in\K}\langle\y,\x\rangle\] which further implies that
\[L(\u)=\E_{\v\in\B}\left(\nabla\mathcal{M}_{\K}\left(\u+\frac{\v}{\delta}\right)\right)=\nabla\E_{\v\in\B}\left(\mathcal{M}_{\K}\left(\u+\frac{\v}{\delta}\right)\right).\]
Then, by combining with the smoothness of the function $\E_{\v\in\B}\left(\max_{\x\in\K}\left\langle\y+\frac{\v}{\delta},\x\right\rangle\right)$, we have
\[\|L(\u)-L(\x)\|_2\leq\delta nD\left\|\u+\frac{\v}{\delta}-\x-\frac{\v}{\delta}\right\|_2=\delta nD\left\|\u-\x\right\|_2.\]

\subsection{Proof of Lemma \ref{lem2_ospf}}
Note that $\nabla\ell_t(\x)=\nabla_t$ for any $t=1,\dots,T$ and $\x\in\K$. Therefore, it is easy to verify that the decisions $\x_1^\ast,\dots,\x_T^\ast$ are actually the same as those generated by applying Algorithm 3 in \citet{Hazan20} to the loss functions\[\ell_1(\x)=\langle\nabla_1,\x\rangle,\dots,\ell_T(\x)=\langle\nabla_T,\x\rangle.\]
Then, according to the proof of Theorem 10 in \citet{Hazan20}, we have
\[
\sum_{t=1}^T\ell_t(\x_{t}^\ast)-\sum_{t=1}^T\ell_t(\x^\ast)\leq \frac{2D}{\delta}+\frac{\delta nDTG^2}{2}.
\]

\section{Conclusion and Future Work} 
\label{sec5}
In this paper, we take the first step towards understanding the effect of arbitrary delays on projection-free OCO algorithms. To this end, we extend OFW and OSPF, which are the state-of-the-art LO-based projection-free algorithms for OCO with non-smooth and smooth loss functions respectively, into the setting with arbitrary delays. Our analysis demonstrates that compared with regret bounds of the original OFW and OSPF, the proposed delayed variants are robust to a relatively large amount of delay.

An interesting future work is to develop delayed variants for other projection-free online algorithms, especially those that can exploit special properties or other oracles of decision sets to attain nearly the same regret as the projection-based algorithms. For example, in the non-delayed setting, the algorithm proposed by \citet{Zak_SC22} enjoys an $O(\sqrt{T\log T})$ regret bound for convex losses and strongly convex sets. Although one can derive an $O(\sqrt{dT\log T})$ regret bound by combing the algorithm with the technique of \citet{Joulani13}, it is more appealing to develop an improved  variant with an $O(\sqrt{\bar{d}T\log T})$ regret bound. 

Moreover, it is worthy to notice that similar to OFW in the non-delayed setting, our delayed OFW can also exploit the strong convexity of decision sets to improve the regret for convex and strongly convex losses to $O(T^{2/3}+\bar{d}T^{1/3})$ and $O(\sqrt{T}+d\log T)$, respectively (see Appendix \ref{sec-app1} for details). Therefore, even the $O(\sqrt{\bar{d}T\log T})$ regret is worst than our $O(\sqrt{T}+d\log T)$ regret for strongly convex losses if $d$ does not exceed $O(\sqrt{\bar{d}T/\log T})$. We will investigate whether our regret bound for strongly convex losses and decision sets can be further improved.





\appendix
\section{Additional Results of Delayed OFW}
\label{sec-app1}
In the main paper, we only consider delayed OCO over general convex sets. Here, we demonstrate that similar to OFW in non-delayed setting, our delayed OFW can enjoy better regret bounds for convex and strongly convex losses if sets are strongly convex. 
\subsection{Convex Losses}
First, we introduce the definition for strongly convex sets \citep{Gaber_ICML_15,FW-ICML21}.
\begin{myDef}
\label{def3}
A convex set $\K\in\mathbb{R}^n$ is called $\beta_K$-strongly convex if it holds that
\[
\gamma\x+(1-\gamma)\y+\gamma(1-\gamma)\frac{\beta_K}{2}\|\x-\y\|_2^2\z \in\K
\]
for any $\x,\y\in\K$, $\gamma\in[0,1]$, and $\z\in\mathbb{R}^n$ with $\|\z\|_2=1$.
\end{myDef}
Then, we establish the following theorem regarding the regret of delayed OFW for convex losses and strongly convex sets.
\begin{thm}
\label{thm0-sc}
Suppose the decision set is $\beta_K$-strongly convex, and Assumptions \ref{assum1} and \ref{assum2} hold. For any $\x^\ast\in\K$, Algorithm \ref{alg1} with $\eta=\frac{D}{2GT^{2/3}}$ has
\[\sum_{t=1}^Tf_t(\x_t) - \sum_{t=1}^Tf_t(\x^\ast)\leq G(3\sqrt{\gamma}+2D)T^{2/3}+\frac{GD(1+\bar{d})T^{1/3}}{2}\]
where $\gamma=\max(4D^2,64/(\beta_K^2))$.
\end{thm} 
\begin{proof}
This proof is similar to that of Theorem \ref{thm0}, and we still need to use (\ref{thm0_eq1}) and Lemma \ref{c_c_lemma_term_A}. The main difference is that Lemmas \ref{thm0_lem1} and \ref{thm0_lem0} can be improved by utilizing the strong convexity of decision sets. Specifically, let $\y_{t}^\ast=\argmin_{\y\in \K}F_{t-1}(\y)$ for any $t\in[T+1]$, where $F_t(\y)$ is defined in (\ref{sec4-1-eq1}). We first prove that under assumptions used in Theorem \ref{thm0-sc}, Algorithm \ref{alg1} with $\eta=\frac{D}{2GT^{2/3}}$ has
\begin{equation}
\label{eq1-app}
F_{t-1}(\y_{t})-F_{t-1}(\y_t^\ast)\leq\frac{\gamma}{T^{2/3}}
\end{equation}
for any $t\in[T+1]$, which has tighter dependence on $T$ than the upper bound in Lemma \ref{thm0_lem1}.

For brevity, we define $h_{t}=F_{t-1}(\y_{t})-F_{t-1}(\y_t^\ast)$ for $t\in[T+1]$ and $h_t(\y_{t-1})=F_{t-1}(\y_{t-1})-F_{t-1}(\y_t^\ast)$ for $t=2,\dots,T+1$. For $t=1$, since $\y_1=\argmin_{\y\in\K}\|\y-\y_1\|_2^2$, we have
\begin{equation}
\label{thm0_lem1-sc_eq1}
h_1=F_{0}(\y_1)-F_{0}(\y_1^\ast)=0\leq\frac{\gamma}{T^{2/3}}.
\end{equation}
Then, for any $T+1\geq t\geq2$, we notice that (\ref{thm0_lem1_eq2}) still holds. Therefore, by assuming $h_{t-1}\leq\frac{\gamma}{T^{2/3}}$ and setting $\eta=\frac{D}{2GT^{2/3}}$, we have
\begin{equation}
\label{thm0_lem1-sc_eq2}
\begin{split}
h_t(\y_{t-1})\leq& h_{t-1}+\eta G\sqrt{h_{t-1}}+\eta^2G^2\\
\leq&\frac{\gamma}{T^{2/3}}+\frac{D\sqrt{\gamma}}{2T}+\frac{D^2}{4T^{4/3}}\\
\leq&\frac{\gamma}{T^{2/3}}+\frac{\gamma}{4T}+\frac{\gamma}{16T^{4/3}}\\
\leq&\frac{\gamma}{T^{2/3}}\left(1+\frac{1}{2T^{1/3}}\right)
\end{split}
\end{equation}
where the third inequality is due to $4D^2\leq\gamma$.

To further bound $h_t$, we introducing the following lemma.
\begin{lem}
\label{lem_SC_sets}(Derived from Lemma 1 of \citet{Gaber_ICML_15})
Let $f(\x):\K\to\R$ be a convex and $\alpha$-smooth function, where $\K$ is $\beta_K$-strongly convex. Moreover, let $\x_{\ii}\in\K$ and $\x_{\oo}=\x_{\ii}+\sigma^\prime(\v-\x_{\ii})$, where $\v\in\argmin_{\mathbf{x}\in\mathcal{K}} \langle\nabla f(\x_{\ii}),\mathbf{x}\rangle$ and $\sigma^\prime=\argmin_{\sigma\in[0,1]}\langle\sigma(\v-\x_{\ii}),\nabla f(\x_{\ii})\rangle+\frac{\sigma^2\alpha}{2}\|\v-\x_{\ii}\|_2^2$. For any $\x^\ast\in\argmin_{\x\in \K}f(\x)$, we have
\begin{align*}
f(\x_{\oo})-f(\x^\ast)\leq(f(\x_{\ii})-f(\x^\ast))\max\left(\frac{1}{2},1-\frac{\beta_K\|\nabla f(\x_{\ii})\|_2}{8\alpha}\right).
\end{align*}
\end{lem}
Moreover, we notice that when $f(\x):\K\to\mathbb{R}$ is an $\beta$-strongly convex function, \citet{Gaber_ICML_15} have proved that
 \begin{equation}
\label{dual_cor_scvx}
\|\nabla f(\x)\|_2\geq\sqrt{\frac{\alpha}{2}}\sqrt{f(\x)-f(\x_\ast)}
\end{equation}
for $\x^\ast=\argmin_{\x\in \K}f(\x)$ and any $\x\in\K$.

From the definition in (\ref{sec4-1-eq1}), $F_{t-1}(\y)$ is $2$-smooth and $2$-strongly convex for any $t\in[T+1]$. By applying Lemma~\ref{lem_SC_sets} with $f(\x)=F_{t-1}(\y)$ and $\x_{\ii}=\y_{t-1}$, for any $t\in[T+1]$, we have $\x_{\oo}=\y_{t}$ and
\begin{equation}
\label{SC_pro1}
\begin{split}
h_t\leq&h_t(\y_{t-1})\max\left(\frac{1}{2},1-\frac{\beta_K\|\nabla F_{t-1}(\y_{t-1})\|}{16}\right).
\end{split}
\end{equation}
If $\frac{1}{2}\leq\frac{\beta_K\|\nabla F_{t-1}(\y_{t-1})\|}{16}$, by combining (\ref{thm0_lem1-sc_eq2}) with (\ref{SC_pro1}), we have
\begin{equation}
\label{SC_case1}
\begin{split}
h_t\leq&\frac{\gamma}{2T^{2/3}}\left(1+\frac{1}{2T^{1/3}}\right)\leq\frac{\gamma}{T^{2/3}}.
\end{split}
\end{equation}
Otherwise, there still exist two cases. If $h_t(\y_{t-1})\leq\frac{\gamma}{T^{2/3}}$, it is easy to verify that
\begin{equation}
\label{SC_case2}
\begin{split}
h_t\leq h_t(\y_{t-1})\leq\frac{\gamma}{T^{2/3}}.
\end{split}
\end{equation}
If $h_t(\y_{t-1})>\frac{\gamma}{T^{2/3}}$, we have 
\begin{equation}
\label{SC_case3}
\begin{split}
h_t\leq h_t(\y_{t-1})\leq&\frac{\gamma}{T^{2/3}}\left(1+\frac{1}{2T^{1/3}}\right)\left(1-\frac{\beta_K\|\nabla F_{t-1}(\y_{t-1})\|}{16}\right)\\
\leq&\frac{\gamma}{T^{2/3}}\left(1+\frac{1}{2T^{1/3}}\right)\left(1-\frac{\beta_K\sqrt{h_t(\y_{t-1})}}{16}\right)\\
\leq&\frac{\gamma}{T^{2/3}}\left(1+\frac{1}{2T^{1/3}}\right)\left(1-\frac{\beta_K\sqrt{\gamma}}{16T^{1/3}}\right)\\
\leq&\frac{\gamma}{T^{2/3}}\left(1+\frac{1}{2T^{1/3}}\right)\left(1-\frac{1}{2T^{1/3}}\right)\leq\frac{\gamma}{T^{2/3}}
\end{split}
\end{equation}
where the second inequality is due to (\ref{dual_cor_scvx}), and the fourth inequality is due to $\gamma\geq64/(\beta_K^2)$.

By combining (\ref{thm0_lem1-sc_eq1}), (\ref{SC_case1}), (\ref{SC_case2}), and (\ref{SC_case3}), we have proved (\ref{eq1-app}) for any $t\in[T+1]$. Then, by combining $\eta=\frac{D}{2GT^{2/3}}$ and (\ref{eq1-app})  with Lemma \ref{c_c_lemma_term_A}, we have
\begin{equation}
\label{eq2-app}
\begin{split}
\sum_{t=1}^TG\|\y_{\tau_{c_t}}-\y_{t}\|_2
\leq&\frac{GD\bar{d}T^{1/3}}{2}+2\sqrt{\gamma}GT^{2/3}.
\end{split}
\end{equation}
Moreover, we notice that (\ref{thm0_lem0_eqfinal}) still holds here. Therefore, by combining (\ref{eq1-app}) and $\eta=\frac{D}{2GT^{2/3}}$ with (\ref{thm0_lem0_eqfinal}), we have
\begin{equation}
\label{eq3-app}
\begin{split}
\sum_{t=1}^{T}\langle\nabla f_{c_t}(\x_{c_t}),\y_{t}-\x^\ast\rangle\leq&\sqrt{\gamma} GT^{2/3}+2GDT^{2/3}+\frac{GDT^{1/3}}{2}
\end{split}
\end{equation}
which has tighter dependence on $T$ than the upper bound in Lemma \ref{thm0_lem0}.

Finally, we complete this proof by substituting (\ref{eq2-app}) and (\ref{eq3-app}) into (\ref{thm0_eq1}).

\end{proof}
\begin{remark}
\emph{Theorem \ref{thm0-sc} shows that if the set is strongly convex, our Algorithm \ref{alg1} can achieve an $O(T^{2/3}+\bar{d}T^{1/3})$ regret bound for convex losses with arbitrary delays, which is better than the $O(T^{3/4}+\bar{d}T^{1/4})$ regret bound presented in Theorem \ref{thm0}, when the term involving $\bar{d}$ is not dominant. Moreover, we notice that in the non-delayed setting, the projection-free algorithm in \citet{Zak_SC22} actually can achieve an $O(\sqrt{dT\log T})$ regret bound for convex losses and strongly convex sets, which is better than that of OFW. Thus, it is natural to investigate whether the algorithm in \citet{Zak_SC22} can be extended to improve our $O(T^{2/3}+\bar{d}T^{1/3})$ bound in the delayed setting. One straightforward way is to derive an $O(\sqrt{dT\log T})$ regret bound by combining the technique of \citet{Joulani13} with the algorithm in \citet{Zak_SC22}, which is better than our bound when $d$ dose not exceed $O(T^{1/3})$ or $d\approx\bar{d}$. However, if $d=\Omega(T^{1/3})$ and $\bar{d}$ does not exceed $O(T^{1/3})$, this result is worse than our bound. To address this limitation, one potential way is to develop an improved variant of the algorithm in \citet{Zak_SC22} to establish the $O(\sqrt{\bar{d}T\log T})$ regret bound, which has been left as a future work.
}
\end{remark}
\subsection{Strongly Convex Losses}
Furthermore, we consider strongly convex losses, and establish the following theorem.
\begin{thm}
\label{thm1-sc}
Suppose Assumptions \ref{assum1} and \ref{assum2} hold, all losses are $\beta$-strongly convex, and the decision set is $\beta_K$-strongly convex. For any $\x^\ast\in\K$, Algorithm \ref{alg2} has
\begin{align*}
\sum_{t=1}^Tf_t(\x_t)-\sum_{t=1}^Tf_t(\x^\ast)\leq&(C_1+C_2)\left(3{d}D+{d}\sqrt{\frac{2\gamma}{\beta}}+2\sqrt{\frac{2\gamma T}{\beta}}+\frac{2{d}C_1}{\beta}\ln T\right)\\
&+\gamma\sqrt{2T}+\frac{\gamma\ln T}{2}+C_1D
\end{align*}
where $\gamma=\max\left(\frac{4(G+2\beta D)^2}{\beta},\frac{288\beta}{\beta_K^2}\right)$, $C_1=G+\beta D$, and $C_2=G+2\beta D$.
\end{thm}
\begin{proof}
This proof is similar to that of Theorem \ref{thm1}, and we still need to use (\ref{thm1_eq4}) and Lemma \ref{lem_sc_sum_terms}. The main difference is that Lemmas \ref{thm1_SC_OFW1} and \ref{thm1_SC_OFW2} can be improved by utilizing the strong convexity of decision sets. 

First, as discussed in the proof of Lemmas \ref{thm1_SC_OFW1} and \ref{thm1_SC_OFW2}, decisions $\y_1,\dots,\y_{T+1}$ in our Algorithm \ref{alg2} are actually generated by performing OFW for strongly convex losses (see Algorithm 2 in \citet{Wan-AAAI-2021-C} for details) on functions $\tilde{f}_1(\y),\dots,\tilde{f}_T(\y)$, where $
\tilde{f}_t(\y)=\langle \nabla f_{c_t}(\x_{c_t}),\y\rangle+\frac{\beta}{2}\|\y-\y_t\|_2^2$ for $t\in[T]$ is $(G+\beta D)$-Lipschitz and $\beta$-strongly convex. As a result, we can derive some useful results from existing theoretical guarantees of OFW for strongly convex losses.

Specifically, let $\y_{t}^\ast=\argmin_{\y\in \K}F_{t-1}(\y)$ for any $t=2,\dots,T+1$, where $F_{t}(\y)$ is defined in (\ref{sec4-1-eq2}). Because of $F_\tau(\y)=\sum_{i=1}^\tau\tilde{f}_i(\y)$, by applying Lemma 2 of \citet{Wan-AAAI-2021-C} with functions $\tilde{f}_1(\y),\dots,\tilde{f}_T(\y)$, it is not hard to verify that
\begin{equation}
\label{eq4-app}
F_{t-1}(\y_{t})-F_{t-1}(\y_t^\ast)\leq\gamma
\end{equation}
for any $t=2,\dots,T+1$.

Moreover, by applying Theorem 2 of \citet{Wan-AAAI-2021-C} functions $\tilde{f}_1(\y),\dots,\tilde{f}_T(\y)$, we have
\begin{equation}
\label{wan_sc_sc}
\begin{split}
\sum_{t=1}^{T}\left(\langle\nabla f_{c_t}(\x_{c_t}),\y_{t}-\x^\ast\rangle-\frac{\beta}{2}\|\y_t-\x^\ast\|_2^2\right)\leq& \gamma\sqrt{2T}+\frac{\gamma\ln T}{2}+C_1D.
\end{split}
\end{equation}
Then, by combining Lemma \ref{lem_sc_sum_terms} with (\ref{eq4-app}), we have
\begin{equation}
\label{final_eq1-sc}
\begin{split}
\sum_{t=1}^T\|\y_{\tau_{c_t}}-\y_{t}\|_2\leq 6{d}D+2{d}\sqrt{\frac{2\gamma}{\beta}}+4\sqrt{\frac{2\gamma T}{\beta}}+\frac{4{d}C_1}{\beta}\ln T.
\end{split}
\end{equation}
Finally, by substituting (\ref{wan_sc_sc}) and (\ref{final_eq1-sc}) into (\ref{thm1_eq4}), we complete this proof.
\end{proof}
\begin{remark}
\emph{Theorem \ref{thm1-sc} shows that if the set is strongly convex, our Algorithm \ref{alg2} can achieve an $O(\sqrt{T}+{d}\log T)$ regret bound for strongly convex losses with arbitrary delays. This bound is better than the $O(T^{2/3}+{d}\log T)$ regret bound presented in Theorem \ref{thm1}, which only utilizes the convexity condition of sets. Moreover, as long as ${d}$ does not exceed $O(\sqrt{T}/\log T)$, this bound matches the $O(\sqrt{T})$ regret bound of OFW for strongly convex losses and sets in the non-delayed setting \citep{Wan-AAAI-2021-C}, and thus is better than the $O(T^{2/3}+\bar{d}T^{1/3})$ regret bound in Theorem \ref{thm0-sc}, which only utilizes the convexity of functions. Finally, if ${d}$ does not exceed $O(T/\log T)$, this bound is better than the $O(\sqrt{dT\log T})$ regret bound derived by combining the technique of \citet{Joulani13} with the projection-free algorithm in \citet{Zak_SC22}.
}
\end{remark}


\vskip 0.2in
\bibliography{ref}

\end{document}